\documentclass[journal]{IEEEtran}
\usepackage{amsmath,amsfonts}
\usepackage{array}
\usepackage{textcomp}

\usepackage{stfloats}
\usepackage{placeins}
\usepackage{url}
\usepackage{verbatim}

\usepackage{graphicx}
\usepackage{hyperref}  % 支持超链接引用
\usepackage{cleveref}  % 增强\autoref功能，推荐替代原生\ref

\usepackage[table]{xcolor} %gxd
\usepackage{amssymb}
\usepackage{multirow}
\usepackage{pifont}
\usepackage{pgffor}      % 用于实现循环
\usepackage{makecell}
\usepackage{enumitem}
\usepackage{algorithm}
\usepackage{algorithmic}
\usepackage{booktabs}       % professional-quality tables
\usepackage{nicefrac}    % compact symbols for 1/2, etc.
\usepackage{microtype}      % microtypography
\usepackage{adjustbox} %用于figure5垂直注释居中

\def\etal{\emph{et al.}}

\definecolor{BlueGreen}{HTML}{008080}
\definecolor{OliveGreen}{RGB}{128, 128, 0}

\definecolor{mydarkdarkgreen}{RGB}{93, 150, 74} % 用于字体颜色标记 & 用于打勾
\definecolor{mydarkgreen}{RGB}{216, 233, 199}  % 段首颜色标记
\definecolor{mylightgreen}{RGB}{245, 249, 241}

\definecolor{RedOrange}{RGB}{255, 165, 0}

\definecolor{lightgray}{gray}{.9}
\newcolumntype{I}{!{\vrule width 1pt}}
\makeatletter
\newcommand{\thickhline}{%
    \noalign {\ifnum 0=`}\fi \hrule height 1pt
    \futurelet \reserved@a \@xhline
}
\makeatother

\begin{document}

%%%%%%%%% TITLE

% \title{Stereo Factory: Merging Model for Unifying Stereo Matching}
\title{StereoFactory: A Unified Merging Framework for Robust Stereo Matching}
% A Training-Free Solution to Robust Stereo Matching in Model Merging

\author{\normalsize{
Xianda Guo, Pinhan Fu, Ruilin Wang, Wenke Huang, Mang Ye, Qin Zou

\IEEEcompsocitemizethanks{
\IEEEcompsocthanksitem Xianda Guo is with the School of Computer Science, Wuhan University, Wuhan, China, and also with D-Star Robotics, Beijing, China.
E-mail: \href{mailto:xianda_guo@163.com}{xianda\_guo@163.com}
\IEEEcompsocthanksitem Pinhan Fu, Mang Ye, and Qin Zou are with the School of Computer Science, Wuhan University, Wuhan, China.
\IEEEcompsocthanksitem Ruilin Wang is with the Institute of Automation, Chinese Academy of Sciences, Beijing, China, and also with D-Star Robotics, Beijing, China.
\IEEEcompsocthanksitem Wenke Huang is with the College of Computing and Data Science, Nanyang Technological University, Singapore. 
\IEEEcompsocthanksitem Corresponding authors: Mang Ye and Qin Zou. 
% E-mail: \href{mailto:mangye16@gmail.com}{yemang@whu.edu.cn}; \href{mailto:qzou@whu.edu.cn}{qzou@whu.edu.cn}
}}}

\maketitle

\begin{abstract}
Stereo matching has advanced through foundation models trained on large-scale datasets, yet this paradigm suffers from a scalability bottleneck: incorporating new data requires costly joint retraining. Model merging offers a scalable post-hoc alternative by integrating knowledge from specialized models after source checkpoints are available. However, existing merging methods typically retain all available models or rely on greedy inclusion, which can preserve harmful task-vector interference. We propose StereoFactory, a coarse-to-fine evolutionary framework for adaptive model merging. Stage~1 employs a genetic algorithm to search the combinatorial space of model subsets, determining \textit{which} models should participate. Stage~2 addresses module-level knowledge specialization (different functional modules exhibit distinct preferences for knowledge sources) through CMA-ES optimization of architecture-adaptive routing over the selected task vectors, with optional module-level scaling. Experiments across two architectures and four benchmarks demonstrate that StereoFactory consistently achieves the best four-benchmark average under the same checkpoint pool, reducing the average error from 3.80 to 3.30 on NMRF and from 2.88 to 2.19 on FoundationStereo relative to the strongest controlled baseline. The post-hoc search requires only 2.7--3.7\% of the corresponding joint-retraining wall-clock time. Analysis reveals that knowledge contributions are inherently module-specific, and selected subsets can transfer across architectures with minimal degradation.
Code will be publicly released upon acceptance at: \url{https://github.com/XiandaGuo/StereoFactory}.
\end{abstract}


\section{Introduction}
\label{sec:intro}

\IEEEPARstart{S}{tereo} matching estimates dense depth maps by establishing pixel-wise correspondences
between rectified image pairs, serving as a fundamental component in autonomous
driving~\cite{kitti2012}, robotics~\cite{khazatsky2024droid}, and 3D
reconstruction systems~\cite{xu2020aanet,ganetADL}.
Recent stereo systems, especially foundation-style models trained on broad
data mixtures~\cite{wen2025foundationstereo}, have
achieved remarkable zero-shot generalization through training on aggregated
datasets. However, this training paradigm exhibits poor scalability:
incorporating new datasets necessitates complete retraining on the union of all
data sources, incurring substantial computational overhead that scales with the
total data volume.

\begin{figure}[t]
    \centering
    \includegraphics[width=\columnwidth]{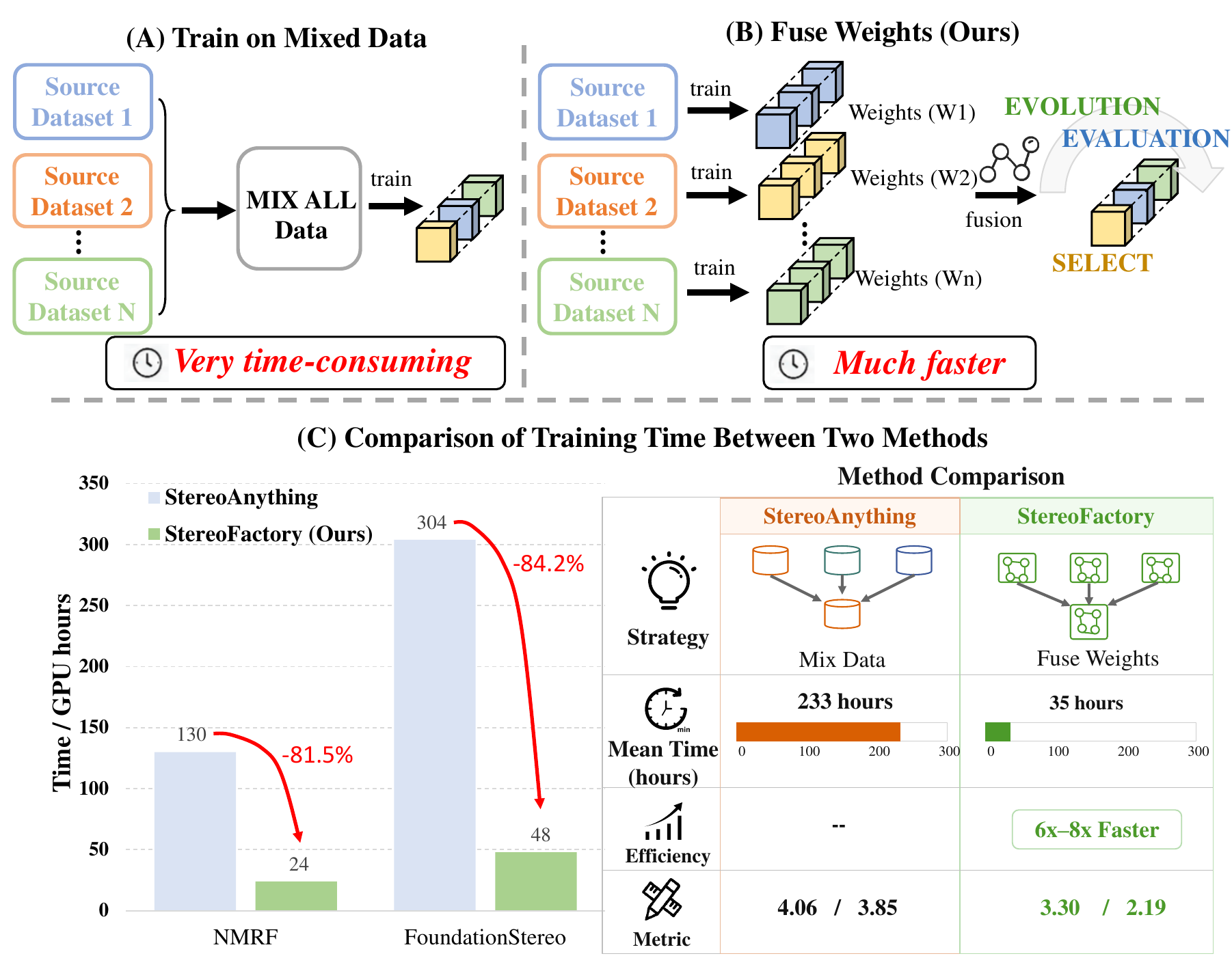}
    \vspace{-20pt}
\caption{Motivation of StereoFactory. Lower is better for all metrics. Conventional mixed-data training retrains a
single model on the union of all source datasets whenever new data sources are
introduced. StereoFactory instead keeps independently trained source-specific
checkpoints reusable and updates only the post-hoc merging stage through
evolutionary selection and adaptive composition.}
    \label{fig:teaser}
\end{figure}

Model merging~\cite{ilharco2023editing,yadav2023ties,yang2024adamerging,wortsman2022model}
presents a compelling alternative by combining independently trained models in weight
space without joint retraining. This paradigm enables efficient knowledge
integration while circumventing the computational burden of training a single
model on the union of all source data.
Nevertheless, many merge-all approaches combine all available models with equal
or similarity-based weights, implicitly assuming that incorporating more models
will not introduce harmful interference.

We challenge this assumption empirically. Fig.~\ref{fig:teaser} summarizes the
paradigm shift motivating our design: instead of either retraining on all source
data or uniformly averaging every available checkpoint, StereoFactory composes a
selective set of useful source models. Our controlled experiments later verify
that merging all eleven dataset-specific models can be suboptimal, whereas
selectively merging a smaller subset achieves stronger average generalization. This
counter-intuitive finding indicates that certain models introduce conflicting
representations that degrade generalization, underscoring the necessity of
adaptive model selection for effective merging.

The key question then becomes: \textit{which} models should be merged, and how can
we efficiently identify them? Existing selection approaches, exemplified by the
greedy strategy in Model Soups~\cite{wortsman2022model}, iteratively add models that
improve validation performance. While intuitive, greedy search is inherently
sequential and susceptible to local optima: an early inclusion decision cannot be
revisited, and the final subset depends heavily on the evaluation order. Moreover,
the search space grows combinatorially. For $N{=}11$ datasets, there are
$\sum_{k=2}^{11}\binom{11}{k} = 2036$ possible subsets of size two or more, which makes exhaustive
evaluation intractable and demands a more principled search strategy.

Beyond the question of \textit{which} models to merge, a subtler issue arises
regarding the \textit{granularity} at which merging operates. Conventional methods
apply a single set of merging coefficients globally across all network parameters,
treating the entire model as a monolithic entity. However, modern stereo architectures
comprise functionally distinct modules: feature extraction backbones, cost
aggregation networks, and disparity refinement heads, each responsible for different
aspects of the matching pipeline. We hypothesize, and subsequently verify through
systematic probing experiments, that these modules exhibit heterogeneous preferences
for knowledge sources: the best source model for one module may differ substantially
from that of another. We refer to this phenomenon as \textit{module-level knowledge
specialization}, which implies that global merging strategies are limited in their
capacity to exploit the structured complementarity among specialized models.

Motivated by these observations, we propose StereoFactory, a coarse-to-fine
evolutionary framework for adaptive model merging in stereo matching. StereoFactory
operates in two stages of increasing granularity. In the first stage, a genetic
algorithm searches the combinatorial space of model subsets to determine which
specialized models should participate in merging. The evolutionary search naturally
handles variable-length discrete representations, balances exploration and
exploitation, and circumvents the limitation of irrevocable early decisions inherent
in greedy approaches. The task vectors from the best selected subset provide the
selected pool for the second stage.
In the second stage, StereoFactory performs architecture-adaptive routing over
the selected task vectors. Rather than assigning a single global
weight, a CMA-ES optimizer~\cite{hansen2016cma} searches for block-specific
routing weights and optional local structural scaling factors. The covariance
adaptation mechanism in CMA-ES is well-suited to this setting, as
it automatically captures interactions between candidate routing and module scales,
allowing different modules to favor different knowledge sources and adaptation
strengths.
This two-stage pipeline progressively transitions from discrete subset selection to
continuous structural optimization, addressing the \textit{what to merge} and
\textit{how to merge} questions at appropriate levels of granularity. The entire
search process requires only a lightweight validation split and incurs modest
cost compared to retraining, as each evaluation involves only model inference rather
than gradient computation.

We conduct comprehensive experiments across two architectures spanning diverse
design paradigms: NMRF~\cite{NMRFStereo}, which formulates stereo matching as
neural Markov random field inference; and
FoundationStereo~\cite{wen2025foundationstereo}, a large-scale foundation model
with transformer components. These are evaluated on four standard benchmarks:
ETH3D~\cite{schops2017eth3d}, KITTI 2012~\cite{kitti2012},
KITTI 2015~\cite{menze2015kitti15}, and
Middlebury~\cite{scharstein2014middlebury}. Results demonstrate that
StereoFactory consistently achieves the best four-benchmark average over both
uniform averaging and widely used
interference-reduction methods, including TIES-Merging~\cite{yadav2023ties} and
DARE~\cite{yu2024dare}, under the same checkpoint pool. We further show that the
post-hoc search costs only 2.7--3.7\% of joint retraining time after source
checkpoints are available, and that the
second-stage module-level routing optimization yields additional gains beyond subset
selection alone, confirming that different architectural modules benefit from
distinct knowledge sources. Reusable-subset experiments further show that selected
source subsets can transfer across architectures with minimal degradation,
supporting our central thesis that robust stereo merging is a selection-and-routing
problem rather than merely a global averaging problem.

\vspace{1mm}
Our contributions are summarized as follows:
\begin{itemize}[leftmargin=*,itemsep=2pt,topsep=2pt]
    \item We empirically demonstrate that adaptive model selection is a central
    factor for effective stereo model merging: selectively merging a compact subset
    of models consistently achieves the best four-benchmark average over the
    prevalent merge-all paradigm as well as widely used interference-reduction
    methods across two architectures.

    \item We identify the phenomenon of module-level knowledge specialization in
    merged stereo networks, where different functional modules exhibit distinct
    preferences for knowledge sources, and propose a coarse-to-fine framework that
    exploits this structure through CMA-ES-based architecture-adaptive routing
    over selected task vectors.

    \item We introduce StereoFactory, a two-stage evolutionary framework that
    efficiently navigates both the discrete model selection space and the continuous
    module-level routing space, requiring only a lightweight validation split
    and no joint retraining after source checkpoints are available.

    \item We provide systematic analysis revealing that knowledge contributions
    are inherently module-specific, and that selected subsets can transfer across
    architectures with minimal degradation, offering insights into dataset
    complementarity for stereo model merging.
\end{itemize}
\section{Related Work}
\label{sec:related}

\subsection{Deep Stereo Matching}
Deep stereo matching has evolved from end-to-end cost-volume
networks~\cite{sceneflow,gcnet,psmnet2018,gwcnet2019,gao2026spikestereonetbraininspiredframeworkstereo,guan2026iclr-dispvit,wang2026ec} to recurrent refinement
and stronger cost filtering~\cite{lipson2021raft,li2022crestereo,igev2023,NMRFStereo,Zhang2026GeometryAwareSM,wen2026fastfoundationstereorealtimezeroshotstereo}.
Recent stereo matching studies~\cite{Yang2021RLStereo,Deng2021detailStereo,
Zeng2022HysteresisStereo,Wang2025ADStereo,Dai2025MGSStereo,Gray2025MDE,Wang2026SMFormer,Cheng2022TwoBranchStereo,zheng2026pipstereoprogressiveiterationspruner} have further improved stereo systems from various aspects, including efficiency, matching accuracy, multi-view estimation, and self-supervised learning.
Motivated by the success of MiDaS~\cite{midas} and DepthAnything~\cite{depthanytingv1, depthanytingv2}, StereoAnything~\cite{guo2024stereo} has demonstrated that scaling up training data can substantially improve the cross-domain generalization capability of stereo matching models. However, this training paradigm typically requires expensive joint retraining whenever new source datasets become available, which limits scalability and practical maintainability. This limitation motivates the need for a unified post-hoc model merging framework that can effectively reuse source-specific checkpoints and adaptively integrate their complementary knowledge without retraining from scratch.
% Despite these advances, cross-domain generalization remains difficult because
% models trained on synthetic or limited source domains often degrade on real-world
% scenes. Recent work improves robustness through domain-invariant
% learning~\cite{dsmnet2020,wang2024dust3r}, masked representation
% learning~\cite{rao2023masked}, and large-scale foundation-style training~\cite{guo2024stereo,wen2025foundationstereo}. 
% These strategies improve zero-shot performance, but usually require expensive joint retraining when new source datasets are introduced, motivating a post-hoc alternative that reuses source-specific checkpoints.

\subsection{Model Merging}
Model merging combines multiple task-specific models into a unified model, often
avoiding full joint retraining and offering a cost-effective alternative to joint
learning~\cite{yang2024model,yadav2024survey}. Representative approaches include
weight averaging and Model Soups~\cite{wortsman2022model}, task-vector
arithmetic~\cite{ilharco2023editing}, and interference-reduction methods such as
TIES-Merging~\cite{yadav2023ties} and DARE~\cite{yu2024dare}. More adaptive
methods learn merging coefficients directly: AdaMerging~\cite{yang2024adamerging}
uses entropy minimization on unlabeled target-domain data, while
Fisher Merging~\cite{matena2022merging} weights parameters using Fisher
information.
Most existing methods, however, assume a fixed pool of candidate checkpoints and
focus on \emph{how} to combine all available task vectors. The preceding question
of \emph{which} source checkpoints should participate is less explored. Model
Soups provides a greedy selection mechanism, but its sequential decisions cannot
be revised, and it operates at a global granularity without modeling module-level
knowledge specialization. Recent evolutionary merging recipes~\cite{akiba2025evolutionary}
demonstrate the promise of search-based merging, but primarily optimize global or
layer-wise weights in a predefined pool. StereoFactory instead treats subset
selection as a primary optimization target and then performs block-wise adaptive
routing for dense stereo matching.
\subsection{Evolutionary Computation in Deep Learning}
Evolutionary algorithms have been widely used for neural architecture search and
hyperparameter optimization~\cite{liu2023enas_survey,loshchilov2016cmaes,back2023evolutionary,li2024evol_survey,bi2023evolutionary_cv}. They are attractive for black-box objectives with discrete choices or
non-differentiable metrics because they require only fitness evaluations.
CMA-ES~\cite{hansen2016cma} is well-suited for continuous black-box
optimization because it adapts a covariance structure without requiring
gradients. This matches our model-merging setting: a genetic algorithm handles
the variable-length source-subset search, and CMA-ES optimizes continuous
block-wise routing weights in the second stage.
% ============================================
% SECTION 3: Preliminaries
% ============================================

\section{Preliminaries}
\label{sec:preliminaries}

\subsection{Problem Setup}

We consider a collection of $N$ labeled stereo datasets $\{\mathcal{D}_i\}_{i=1}^{N}$,
where each $\mathcal{D}_i = \{(I_L^k, I_R^k, d^k)\}_{k=1}^{n_i}$ contains $n_i$
rectified stereo pairs with ground-truth disparity maps, spanning diverse domains
including synthetic rendering, indoor environments, outdoor driving, and robotic
manipulation. Let $f_\theta$ denote a stereo matching network parameterized by
$\theta$. Given a pre-trained base model $\theta_{\text{base}}$ and $N$ training
datasets, our goal is to obtain a merged model $\theta_{\text{merged}}$ that
generalizes to target domains without joint retraining.

\subsection{Task Vectors and Model Merging}

Starting from $\theta_{\text{base}}$, we fine-tune on each dataset separately to
obtain $N$ specialized models $\{\theta_i\}_{i=1}^{N}$. Following Task
Arithmetic~\cite{ilharco2023editing}, the task vector for dataset $\mathcal{D}_i$ is:
\begin{equation}
    \tau_i = \theta_i - \theta_{\text{base}},
    \label{eq:task_vector}
\end{equation}
which encodes the knowledge acquired from $\mathcal{D}_i$. The merged model is:
\begin{equation}
    \theta_{\text{merged}} = \theta_{\text{base}} + \sum_{i=1}^{N} w_i \cdot \tau_i,
    \label{eq:merge}
\end{equation}
where $\boldsymbol{w} = (w_1, \ldots, w_N)$ are fusion weights. Existing methods
differ in how they determine $\boldsymbol{w}$: uniform averaging~\cite{wortsman2022model}
sets $w_i = 1/N$; AdaMerging~\cite{yang2024adamerging} learns coefficients via
entropy minimization; TIES-Merging~\cite{yadav2023ties} and DARE~\cite{yu2024dare}
process task vectors to reduce interference. Most of these methods operate on a
fixed merge pool and focus on how to combine the available task vectors. In this
work, we instead ask the preceding question of \emph{which} task vectors should
participate before optimizing how they are composed.

% ============================================
% SECTION 4: StereoFactory
% ============================================

\section{StereoFactory}
\label{sec:method}

\begin{figure*}
    \centering
    \includegraphics[width=1\linewidth]{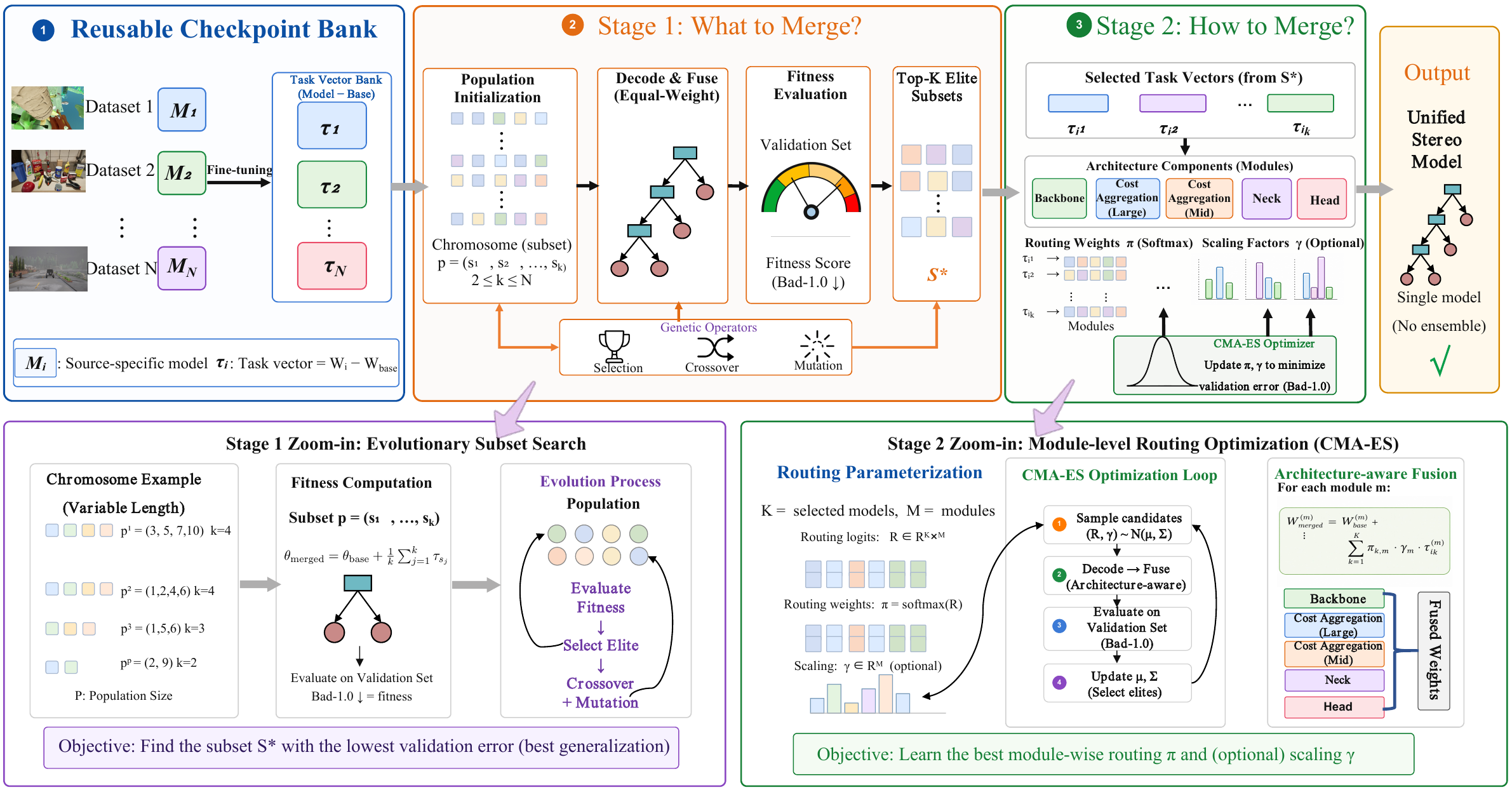}
    \vspace{-20pt}
\caption{Overview of the StereoFactory framework.
Stage~1 searches for the optimal subset of fine-tuned checkpoints via evolutionary
search with validation-set fitness.
Stage~2 further refines the merging through architecture-adaptive routing
across different architectural components.}
    \label{fig:framework}
\end{figure*}

We present StereoFactory, a two-stage coarse-to-fine evolutionary framework that
addresses both \emph{what to merge} and \emph{how to merge} at progressively finer
granularity. As illustrated in Fig.~\ref{fig:framework}, the framework operates as
follows. Stage~1 employs a genetic algorithm~\cite{back2023evolutionary} to search
the combinatorial space of model subsets, identifying \textit{which} specialized
models should participate in merging.
Stage~2 performs module-level routing optimization over the selected task vectors
via CMA-ES~\cite{hansen2016cma}, with optional entropy regularization when sparse
routing is desired. This stage exploits the phenomenon of \textit{module-level
knowledge specialization}: the empirical finding that different functional
modules within a network benefit from distinct knowledge sources.

The key design principle is the decomposition of model merging into two sub-problems
with fundamentally different optimization landscapes. Subset selection is a
combinatorial optimization problem involving variable-length discrete
representations, naturally suited to genetic
algorithms~\cite{li2024evol_survey,bi2023evolutionary_cv}. Module-level weight
allocation is a continuous optimization problem in a structured real-valued space,
where CMA-ES excels due to its ability to adaptively model the local covariance
structure~\cite{loshchilov2016cmaes}. Attempting to solve both simultaneously would
require mixed discrete-continuous optimization over a substantially larger space,
whereas our decomposition keeps each stage tractable.

\subsection{Model Selection Problem}
\label{subsec:formulation}

We seek an optimal subset $\mathcal{S}^* \subseteq \{1, \ldots, N\}$ that
minimizes the validation error on a held-out set $\mathcal{D}_{\text{val}}$:
\begin{equation}
    \mathcal{S}^* =
    \arg\min_{\substack{\mathcal{S}\subseteq\{1,\ldots,N\}\\
    2\leq|\mathcal{S}|\leq N}} \,
    \mathcal{L}_{\text{val}}\!\left(\text{Merge}(\theta_{\text{base}},
    \{\tau_j\}_{j \in \mathcal{S}})\right).
    \label{eq:objective}
\end{equation}
This formulation is challenging for three reasons. First, the search space is
combinatorial: for $N{=}11$ datasets, there are $\sum_{k=2}^{11}\binom{11}{k}
= 2036$ unordered subsets of size two or more, making exhaustive evaluation
intractable. Second, the objective is non-differentiable with respect to
subset membership, precluding gradient-based optimization. Third, greedy approaches
such as Model Soups~\cite{wortsman2022model} are susceptible to local optima:
once a model is included, it cannot be removed, and the final subset depends heavily
on the evaluation order.

\subsection{Model Fusion}
\label{subsec:fusion}

Given a selected subset $\mathcal{S} = \{s_1, s_2, \ldots, s_k\}$, we fuse the
corresponding task vectors through equal-weight averaging:
\begin{equation}
    \theta_{\text{merged}} = \theta_{\text{base}} + \frac{1}{k}\sum_{j=1}^{k} \tau_{s_j},
    \label{eq:equal_avg}
\end{equation}
which treats all selected task vectors symmetrically. This parameter-free scheme
avoids introducing additional hyperparameters during subset search and ensures
that the fitness landscape depends solely on subset composition.

Table~\ref{tab:fusion_compare} compares our approach with existing fusion strategies.

\begin{table}[t]
\centering
\caption{Comparison of model fusion strategies. Our two-stage framework first
selects the optimal subset with equal-weight fusion (Stage~1), then performs
module-level routing optimization via CMA-ES (Stage~2).}
\label{tab:fusion_compare}
\setlength{\tabcolsep}{3pt}
\renewcommand{\arraystretch}{1.05}
\small
\resizebox{\columnwidth}{!}{
\begin{tabular}{lcccc}
\thickhline
\rowcolor[gray]{0.92}
\textbf{Method} & \textbf{Select.} & \textbf{Weight} & \textbf{Gran.} & \textbf{Infer.} \\
\hline\hline
Uniform Avg. & All & Equal ($1/N$) & Global & Single \\
Task Arith.~\cite{ilharco2023editing} & All & Manual $\lambda$ & Global & Single \\
AdaMerging~\cite{yang2024adamerging} & All & Learned & Task/Layer & Single \\
Model Soups~\cite{wortsman2022model} & Greedy & Equal & Global & Single \\
\hline
\rowcolor{cyan!10}
Ours (Stage~1) & EA & Equal ($1/k$) & Global & Single \\
\rowcolor{cyan!10}
Ours (Stage~1+2) & EA+CMA-ES & Adaptive & Module & Single \\
\thickhline
\end{tabular}}
\end{table}

\subsection{Stage~1: Evolutionary Subset Search}
\label{subsec:stage1}

The goal of Stage~1 is to evolve a diverse population of subset candidates and
identify the best-performing selected task-vector pool for Stage~2. We employ a
genetic algorithm that naturally handles variable-length discrete representations
and provides a principled balance between exploration and exploitation.

\subsubsection{Representation}

Each candidate solution is encoded as a variable-length chromosome of
model indices:
\begin{equation}
\begin{aligned}
    p &= (s_1, s_2, \ldots, s_k), \quad s_i \in \{1, \ldots, N\},\\
    s_i &\neq s_j,\quad 2 \leq k \leq N .
\end{aligned}
\label{eq:chromosome}
\end{equation}
where $k$ varies across individuals, allowing the algorithm to simultaneously
explore subsets of different sizes. Each chromosome is decoded into a merged model
via Eq.~\eqref{eq:equal_avg}, and its fitness is defined as the validation error:
\begin{equation}
    f(p) = \text{Bad}_{1.0}(\theta_{\text{merged}}, \mathcal{D}_{\text{val}}).
    \label{eq:fitness}
\end{equation}
We adopt ETH3D as the validation set $\mathcal{D}_{\text{val}}$ due to its coverage
of both indoor and outdoor scenes with sub-pixel accurate ground truth, and we
study alternative validation metrics in Sec.~\ref{subsec:ablation}.

\subsubsection{Genetic Operators}

\textbf{Initialization.} The initial population of $M$ chromosomes is
generated by sampling a subset size $k \sim \text{Uniform}(2, N)$ and
selecting $k$ distinct indices uniformly at random. This ensures diversity in both
composition and size from the outset.

\textbf{Crossover.} Given parents $p^1 = (s^1_1, \ldots, s^1_{k_1})$ and
$p^2 = (s^2_1, \ldots, s^2_{k_2})$, offspring are produced via one-point crossover.
Cut points $c_1 \in [1, k_1{-}1]$ and $c_2 \in [1, k_2{-}1]$ are randomly selected,
and the offspring are formed by exchanging the tails:
\begin{equation}
\begin{aligned}
p^1_o &= (s^1_1, \ldots, s^1_{c_1},\, s^2_{c_2+1}, \ldots, s^2_{k_2}), \\
p^2_o &= (s^2_1, \ldots, s^2_{c_2},\, s^1_{c_1+1}, \ldots, s^1_{k_1}).
\end{aligned}
\label{eq:crossover}
\end{equation}

Duplicate indices within each offspring are removed to maintain the subset
constraint. Offspring containing fewer than 2 elements are augmented with randomly
sampled indices.

\textbf{Mutation.} With probability $p_m$, one of three operators is applied
uniformly at random:
\begin{itemize}[leftmargin=*,itemsep=1pt,topsep=2pt]
    \item \textit{Replace}: substitute the model at a random position with a
          different model index, exploring alternative dataset choices.
    \item \textit{Insert}: add a new model at a random position (if $k < N$),
          expanding the subset to incorporate additional knowledge.
    \item \textit{Delete}: remove a model at a random position (if $k > 2$),
          reducing the subset to eliminate potential interference.
\end{itemize}
These operators collectively enable exploration along two axes: subset
composition (Replace, Insert, Delete) and subset size (Insert, Delete).

\textbf{Selection and Elitism.} We employ binary tournament selection for
parent choice and carry the best individual from each generation directly into the
next, preventing the loss of the current best solution.

\subsubsection{From Population to Selected Pool}

After $T_1$ generations, the best unique individual in the evaluated archive is
selected by validation fitness. Its task vectors form the selected pool:
\begin{equation}
    \mathcal{P}^* = \{\tau_{s}\mid s\in\mathcal{S}^*\},
    \label{eq:selected_pool}
\end{equation}
which serves as the input to Stage~2 in the reported experiments. The evaluated
archive can also retain additional high-fitness subsets for diagnostic analysis,
but the main results use the selected task-vector pool above.

\begin{algorithm}[t]
\caption{Stage~1: Evolutionary Subset Search}
\label{alg:stage1}
\begin{algorithmic}[1]
\REQUIRE Task vectors $\{\tau_i\}_{i=1}^{N}$, base model $\theta_{\text{base}}$,
         validation set $\mathcal{D}_{\text{val}}$, population size $M$,
         generations $T_1$
\ENSURE Selected task-vector pool $\mathcal{P}^*$
\STATE Initialize population $P_0 = \{p_1, \ldots, p_M\}$ randomly
\STATE Evaluate $f(p)$ for all $p \in P_0$
\STATE Initialize evaluated archive $\mathcal{A} \leftarrow P_0$
\FOR{$t = 1$ \TO $T_1$}
    \STATE $Q_t \leftarrow \emptyset$
    \WHILE{$|Q_t| < M$}
        \STATE Select parents $p^a, p^b$ via binary tournament from $P_{t-1}$
        \STATE $(p^a_o, p^b_o) \leftarrow \text{Crossover}(p^a, p^b)$; deduplicate
        \STATE Apply $\text{Mutate}(\cdot)$ to each offspring with prob.\ $p_m$
        \STATE $Q_t \leftarrow Q_t \cup \{p^a_o, p^b_o\}$
    \ENDWHILE
    \STATE Evaluate $f(p)$ for all new $p \in Q_t$ (with caching)
    \STATE $\mathcal{A} \leftarrow \mathcal{A} \cup Q_t$
    \STATE $P_t \leftarrow \text{TournamentSelect}(P_{t-1} \cup Q_t,\, M{-}1)
           \cup \{\text{best}(P_{t-1} \cup Q_t)\}$
\ENDFOR
\STATE Select the best unique individual $\mathcal{S}^*$ from $\mathcal{A}$
\RETURN $\mathcal{P}^* = \{\tau_s \mid s\in\mathcal{S}^*\}$
\end{algorithmic}
\end{algorithm}

\subsection{Stage~2: Architecture-Adaptive Routing via CMA-ES}
\label{subsec:stage2}

Stage~1 yields a compact selected task-vector pool rather than committing to an
all-source merge. For notational uniformity, we denote the selected vectors by
$\{\Delta_e\}_{e=1}^{E}$ and search how each functional block should route among
them. This second stage addresses
\textit{module-level knowledge specialization}: different modules may prefer
different knowledge sources and different adaptation intensities.

\subsubsection{Architecture-Adaptive Search Space}

Let the network parameters be partitioned into $B$ functional blocks
$\{\mathcal{B}_b\}_{b=1}^{B}$ corresponding to semantically coherent
architectural components (e.g., backbone, neck, cost volume, inference head).
The full Stage~2 parameterization consists of block-specific routing logits
$\mathbf{W}\in\mathbb{R}^{B\times E}$ over the candidate task vectors and optional
block scaling factors $\mathbf{s}\in\mathbb{R}^{B}$:
\begin{equation}
    \mathbf{x} = [\mathrm{vec}(\mathbf{W}),\, \mathbf{s}]
    \in \mathbb{R}^{B E + B}.
    \label{eq:cmaes_encoding}
\end{equation}
For each block, the routing logits are mapped to the probability simplex via a
softmax:
\begin{equation}
    \pi_{b,e} = \frac{\exp(W_{b,e})}{\sum_{j=1}^{E}\exp(W_{b,j})},
    \quad e=1,\ldots,E,
    \label{eq:softmax}
\end{equation}
yielding a normalized candidate distribution for each functional block. When
structural intensity modulation is enabled, the scaling factors are bounded with
a scaled sigmoid:
\begin{equation}
    \gamma_b = \gamma_{\min} +
    \frac{\gamma_{\max}-\gamma_{\min}}{1+\exp(-s_b)},
    \quad b=1,\ldots,B,
    \label{eq:sigmoid_scale}
\end{equation}
where $(\gamma_{\min},\gamma_{\max})$ is set according to the stability of the
architecture. In practice we use $[0.25,1.75]$ for the shared-routing
hierarchical variant used in our main experiments.

This formulation contains the architecture-adaptive variants used in our
implementation. Tying the routing weights across blocks
($\pi_{b,e}=\pi_e$) recovers the compact hierarchical fusion used by NMRF and
FoundationStereo, whose search dimension is $E+B$, giving nine dimensions in our
standard NMRF setting. The same routing operator also supports fully block-specific
routing, but the compact tied-routing variant is sufficient for the two
architectures reported in this manuscript.

\subsubsection{Adaptive Routing Fusion}

Given the routing weights $\boldsymbol{\pi}$ and module scales
$\boldsymbol{\gamma}$, we construct the final model block by block. For each
search block $\mathcal{B}_b$, the candidate task vectors are routed first and then
optionally scaled:
\begin{equation}
    \Delta_{\text{final}}^{(\mathcal{B}_b)}
    = \gamma_b \sum_{e=1}^{E} \pi_{b,e}
      \Delta_e^{(\mathcal{B}_b)},
    \label{eq:hierarchical_fusion}
\end{equation}
where $\Delta_e^{(\mathcal{B}_b)}$ denotes the task vector of candidate
$e$ restricted to block $\mathcal{B}_b$. The merged parameters are obtained by
adding this routed task vector back to the base model:
$\theta_{\text{final}}^{(\mathcal{B}_b)}
=\theta_{\text{base}}^{(\mathcal{B}_b)}
+\Delta_{\text{final}}^{(\mathcal{B}_b)}$. Parameters outside the searched
blocks use the first block's routing weights with $\gamma=1.0$, ensuring that
unassigned layers remain aligned with the discovered global routing pattern.

\textbf{Batch Normalization Handling.}
For normalization layers, we decouple activation statistics from task-vector
scaling. Learnable affine parameters participate in Eq.~\eqref{eq:hierarchical_fusion},
but running means and variances are never multiplied by $\gamma_b$. In the full
unified variant, running statistics are routed with the same convex weights
$\pi_{b,e}$ and scale $1.0$. This preserves activation coordinates while still
allowing adaptive knowledge routing, avoiding artificial coordinate shifts when
local module scales are applied.

\subsubsection{CMA-ES Optimization}

We optimize the routing parameters using the Covariance Matrix Adaptation
Evolution Strategy (CMA-ES)~\cite{hansen2016cma}. CMA-ES is derivative-free, which
suits our setting because the fitness, the validation error on ETH3D, is a
non-differentiable benchmark metric. Its covariance adaptation naturally captures
interactions between routing weights and block scales. For sparse expert routing,
we optionally add a small entropy penalty:
\begin{equation}
    \mathcal{F}(\mathbf{x}) =
    \mathrm{Bad}_{1.0}(\theta_{\text{trial}},\mathcal{D}_{\text{val}})
    + \lambda\frac{1}{B}\sum_{b=1}^{B}
      H(\boldsymbol{\pi}_b),
    \label{eq:entropy_fitness}
\end{equation}
where $H(\boldsymbol{\pi}_b)$ is the entropy of block $b$'s candidate routing
distribution. In the reported tied-routing experiments, we set $\lambda=0$
unless explicit sparsification is desired, so the validation metric directly
drives the final routing weights.

\begin{algorithm}[t]
\caption{Stage~2: Architecture-Adaptive Routing via CMA-ES}
\label{alg:stage2}
\begin{algorithmic}[1]
\REQUIRE Selected task-vector pool $\mathcal{P}^*$, base model $\theta_{\text{base}}$, validation set $\mathcal{D}_{\text{val}}$, block partition $\{\mathcal{B}_b\}_{b=1}^{B}$, population $N_{\text{pop}}$, evaluation budget $N_{\text{eval}}$
\ENSURE Final merged model $\theta_{\text{final}}$
\STATE Let $\{\Delta_e\}_{e=1}^{E}$ be the selected task vectors in $\mathcal{P}^*$
\STATE Initialize CMA-ES distribution mean $\mathbf{m}^{(0)} = \mathbf{0}$, step size $\sigma^{(0)}$, and $\mathbf{C}^{(0)} = \mathbf{I}$
\WHILE{evaluation count $< N_{\text{eval}}$ \AND not converged}
    \STATE Sample $\{\mathbf{x}_j\}_{j=1}^{N_{\text{pop}}} \sim \mathcal{N}(\mathbf{m}, \sigma^2 \mathbf{C})$
    \FOR{$j = 1$ \TO $N_{\text{pop}}$}
        \STATE Decode $\mathbf{x}_j$ into block routing weights $\boldsymbol{\pi}$ and module scales $\boldsymbol{\gamma}$ via Eq.~\eqref{eq:softmax} and~\eqref{eq:sigmoid_scale}
        \STATE Construct $\theta_{\text{trial}}$ block-by-block via Eq.~\eqref{eq:hierarchical_fusion}
        \STATE Evaluate fitness $\mathcal{F}(\mathbf{x}_j)$ via Eq.~\eqref{eq:entropy_fitness}
    \ENDFOR
    \STATE Rank solutions and update $\mathbf{m}$, $\sigma$, $\mathbf{C}$ via CMA-ES rules~\cite{hansen2016cma}
\ENDWHILE
\STATE $\mathbf{x}^* \leftarrow \arg\min_{\mathbf{x}} \mathcal{F}(\mathbf{x})$
\RETURN $\theta_{\text{final}}$ decoded from $\mathbf{x}^*$
\end{algorithmic}
\end{algorithm}

\subsection{Complexity Analysis}
\label{subsec:complexity}

\textbf{Stage~1.}
The GA requires $\mathcal{O}(T_1 \times M)$ fitness evaluations, where each
evaluation consists of fusing at most $N$ task vectors followed by one forward pass
on $\mathcal{D}_{\text{val}}$. With parallel evaluation across $W$ GPU workers and
caching of duplicate individuals, the effective wall-clock time is approximately
$T_1 \times \lceil M / W \rceil$ evaluations, minus cache hits.

\textbf{Stage~2 CMA-ES.}
The CMA-ES optimization requires at most $N_{\text{eval}}$ fitness evaluations,
each involving adaptive routing over pre-computed candidate task vectors followed by
one forward pass. The tied-routing hierarchical variant used in the main
experiments reduces to $E+B$ dimensions over the selected dataset task vectors,
yielding a compact nine-dimensional space in our standard NMRF configuration.
All settings remain well within the regime where CMA-ES converges efficiently.

\textbf{Post-hoc search overhead.}
StereoFactory targets the setting where source-specific checkpoints are already
available or can be trained independently. Table~\ref{tab:computational_cost}
therefore reports the additional cost of merging these checkpoints after they are
obtained, and compares it with a conventional joint retraining pass over the same
11 sources. For NMRF, joint retraining requires 48.2 wall-clock hours, whereas our
two-stage search completes in 1.8 hours (3.7\% overhead). The computational
advantage grows with model capacity: FoundationStereo joint retraining requires
128.7 hours, while our search needs only 3.5 hours (2.7\% overhead).
These wall-clock values isolate the post-hoc search stage and are distinct from
the aggregate GPU-hour accounting used for the motivation comparison in
Fig.~\ref{fig:teaser}.
When a new source is added, joint training requires another full union retraining
run; StereoFactory only re-runs the merge search after the new source checkpoint
is available. This creates a reusable checkpoint bank: unchanged source task
vectors can be cached, while the new source contributes only one additional task
vector to the candidate pool. The benefit can therefore compound in incremental
data-collection scenarios, where stereo systems are repeatedly updated with new
synthetic generators, driving scenes, or indoor captures.

\begin{table}[t]
  \centering
  \caption{
    Post-hoc wall-clock search cost comparison on 4$\times$ NVIDIA A100 GPUs.
    The joint retraining row denotes full training on the same 11 sources;
    StereoFactory rows report only the additional merging/search time after
    source-specific checkpoints are available.
  }
  \label{tab:computational_cost}
  \setlength\tabcolsep{4pt}
  \renewcommand\arraystretch{1.15}
  \small
  \resizebox{\columnwidth}{!}{
  \begin{tabular}{l|cc}
    \thickhline
    \rowcolor[gray]{0.92}
    \textbf{Method} & \textbf{NMRF~\cite{NMRFStereo}} & \textbf{FoundationStereo~\cite{wen2025foundationstereo}} \\
    \hline\hline
    Joint retraining (hrs) & 48.2 & 128.7 \\
    \hline
    Stage 1 (hrs)    & 1.2  & 2.4 \\
    Stage 2 (hrs)    & 0.6  & 1.1 \\
    \textbf{Total (hrs)} & \textbf{1.8} & \textbf{3.5} \\
    \hline
    Overhead vs. joint retrain (\%) & 3.7  & 2.7 \\
    \thickhline
  \end{tabular}}
  \vspace{-20pt}
\end{table}

\subsection{Discussion}
\label{subsec:discussion}

\textbf{Parameter-space fusion at all stages.}
A key property of our framework is that both stages produce \emph{single} models.
Stage~1 merges task vectors in parameter space to obtain each individual's model,
and Stage~2 combines candidate task vectors also in parameter space at the module level.
At no point are multiple models required at inference time. This is in contrast to
output-level ensemble methods~\cite{dietterich2000ensemble} that require running
multiple models per input, incurring proportionally higher inference cost.
The ``unified'' aspect of StereoFactory therefore refers to a common search and
routing formulation shared by different stereo backbones; each actual merge is
performed within one weight-compatible architecture initialized from the same base
checkpoint.

\textbf{Relationship to AdaMerging.}
AdaMerging~\cite{yang2024adamerging} also performs layer-wise weight optimization,
but differs from our approach in two fundamental aspects. First, AdaMerging operates
on the original task vectors and requires all models to participate, whereas our
Stage~1 first identifies a beneficial task vector subset, reducing interference before
Stage~2 begins. Second, AdaMerging learns weights via test-time entropy minimization
without labeled data, whereas our CMA-ES directly optimizes the target objective on a
held-out validation set. The two approaches are complementary and could be combined.

\textbf{Relationship to evolutionary model merging.}
Akiba~\etal~\cite{akiba2025evolutionary} also employ evolutionary search for model
merging, but operate exclusively in the continuous weight space without addressing
the discrete subset selection problem. Our two-stage decomposition separates these
fundamentally different optimization landscapes, enabling more effective navigation
of both the combinatorial subset space and the continuous module-level routing space.

\textbf{Why CMA-ES over gradient-based methods?}
The fitness function involves a full forward pass and a non-differentiable
evaluation metric, precluding standard backpropagation. While one could define a
differentiable surrogate loss, this would introduce a discrepancy between the
optimization objective and the actual evaluation criterion. CMA-ES directly
optimizes the true metric, avoiding this mismatch. Moreover, its covariance
adaptation captures correlations among routing and scaling variables without
requiring gradient computation.

\textbf{Limitations.}
The quality of $\theta_{\text{final}}$ depends on the representativeness of
$\mathcal{D}_{\text{val}}$. In our experiments, ETH3D serves as the validation set
due to its scene diversity; therefore, the ETH3D score should be interpreted as
validation-guided rather than strictly unseen-model-selection performance. For
highly specialized target domains, a domain-aligned validation split may yield
better results.
\section{Experiments}
\label{sec:experiments}

% ============================================
% A. Implementation Details
% ============================================
\subsection{Implementation Details}
\label{subsec:impl}

\textbf{Architectures.}
Our experiments are built upon the OpenStereo platform~\cite{guo2023openstereo}.
We evaluate StereoFactory on two stereo matching architectures that span diverse
design paradigms: NMRF~\cite{NMRFStereo}, a neural Markov random field model with
iterative cost filtering; and
FoundationStereo~\cite{wen2025foundationstereo}, a large-scale foundation model
with transformer components. This selection covers structured cost-filtering
and foundation-model regimes, allowing us to test whether the same
merging principle holds across substantially different stereo backbones.

\textbf{Model merging baselines.}
We compare against six representative baselines and model merging strategies, all operating on
the same set of 11 fine-tuned checkpoints per architecture:
(1)~\textit{Single Best}: an oracle single-checkpoint baseline, i.e., the
individual fine-tuned checkpoint with the lowest four-benchmark average for each
architecture;
(2)~\textit{Uniform Averaging}: equal-weight merging of all 11 task vectors;
(3)~\textit{Task Arithmetic}~\cite{ilharco2023editing}: uniform merging with a
tuned scaling factor $\lambda$;
(4)~\textit{Model Soups}~\cite{wortsman2022model}: greedy sequential addition of
checkpoints that improve validation performance;
(5)~\textit{TIES-Merging}~\cite{yadav2023ties}: merging with magnitude-based
trimming and sign conflict resolution;
(6)~\textit{DARE}~\cite{yu2024dare}: random pruning of delta parameters prior to
merging to mitigate interference. AdaMerging~\cite{yang2024adamerging} is
discussed separately because it performs gradient-based entropy minimization on
unlabeled target-domain data, whereas Table~\ref{tab:merging_comparison}
compares gradient-free post-hoc merging methods under the same checkpoint pool.

\textbf{Evolutionary Search.}
The genetic algorithm is configured with a population size of $M{=}50$ over
$T_1{=}15$ generations, with crossover probability $p_c{=}0.8$ and mutation
probability $p_m{=}0.3$. ETH3D~\cite{schops2017eth3d} serves as the validation
set, with Bad-1.0 adopted as the fitness criterion. For our main results, we run
the full search on the lighter-weight NMRF, and for the substantially more
expensive FoundationStereo we adopt the subset discovered on NMRF, directly
instantiating the reusable-subset scenario that motivates our framework. To
quantify the cost of this transfer, we additionally run an architecture-specific
search for each backbone; both directions are reported as an ablation in
Table~\ref{tab:ablation_2}. As shown there, the two configurations differ only
marginally, confirming that selected subsets transfer across architectures with
minimal degradation and that dataset complementarity is not purely
architecture-specific. All experiments are performed on NVIDIA A100 GPUs, with
fitness evaluations within each generation parallelized across available devices.
After the 11 fine-tuned checkpoints are produced, StereoFactory does not revisit
the source training images; merging uses only task vectors, forward-only
validation scores, and no gradient updates.

% ============================================
% B. Datasets and Metrics
% ============================================
\subsection{Datasets and Metrics}
\label{subsec:datasets}

\textbf{Training Datasets.}
For each architecture, we adopt the publicly released checkpoint trained on
SceneFlow~\cite{sceneflow} as the base model $\theta_{\text{base}}$, and
fine-tune it independently on each of $N{=}11$ stereo datasets:
Sintel~\cite{butler2012naturalistic},
FallingThings~\cite{tremblay2018falling},
VirtualKITTI2~\cite{cabon2020virtual},
TartanAir~\cite{wang2020tartanair},
InStereo2K~\cite{bao2020instereo2k},
UnrealStereo4K~\cite{tosi2021smdnets},
CREStereo~\cite{li2022crestereo},
Spring~\cite{mehl2023spring},
DynamicReplica~\cite{karaev2023dynamicstereo},
Carla~\cite{guo2025stereocarla},
and the FoundationStereo dataset~\cite{wen2025foundationstereo}.
All models are fine-tuned using the standard smooth $\ell_1$ loss on disparity
for 10 epochs with a batch size of 8, yielding 22 specialized checkpoints in
total, i.e., 11 per architecture.

\textbf{Evaluation Benchmarks.}
We assess training-free cross-domain generalization on four standard benchmarks
that are held out from the source-specific fine-tuning runs used to build our
checkpoint pool.
ETH3D~\cite{schops2017eth3d} provides high-resolution indoor and outdoor
scenes with sub-pixel accurate ground truth; we report Bad-1.0, defined as the
percentage of pixels with disparity error exceeding 1\,px. 
Following standard practice in model
merging~\cite{wortsman2022model,yang2024adamerging}, ETH3D serves as the
validation set for evolutionary search, using the same Bad-1.0 metric as the
fitness criterion. Crucially, ETH3D is used only for forward-pass fitness
evaluation: no gradient-based optimization, statistic recalibration, or test-time
adaptation is performed on this set. To test whether the search merely overfits
one validation benchmark, Table~\ref{tab:ablation_1} repeats the search with
Middlebury Bad-2.0 and obtains comparable averages (3.21 vs.\ 3.30 on NMRF~\cite{NMRFStereo});
Table~\ref{tab:validation_subset_robustness} further shows consistent selections
when KITTI~15 is used as validation. These checks indicate that the improvements
reflect robust dataset complementarity rather than a single validation metric.
\textbf{KITTI 2012}~\cite{kitti2012} and
\textbf{KITTI 2015}~\cite{menze2015kitti15} are real-world autonomous driving
benchmarks; we report the D1-all metric.
\textbf{Middlebury}~\cite{scharstein2014middlebury} comprises high-resolution
indoor scenes with complex geometry; we report Bad-2.0.

\begin{table*}[t]
\centering
\caption{\textbf{Comparison of model merging strategies on training-free cross-domain generalization.}
Lower is better for all metrics.
All methods share the same 11 checkpoints per architecture.
\textbf{Bold} and \underline{underline} mark the column-wise best and
second-best values within each architecture block.}
\setlength\tabcolsep{4pt}
\renewcommand\arraystretch{1.15}
\resizebox{\textwidth}{!}{
\begin{tabular}{l|ccccc|ccccc}
  \thickhline
  \rowcolor[gray]{0.92}
  & \multicolumn{5}{c|}{\textbf{NMRF}~\cite{NMRFStereo}}
  & \multicolumn{5}{c}{\textbf{FoundationStereo}~\cite{wen2025foundationstereo}} \\
  \rowcolor[gray]{0.92}
  \multirow{-2}{*}{\textbf{Methods}}
  & \textbf{K12} & \textbf{K15} & \textbf{Mid} & \textbf{ETH3D} & \textbf{Mean}
  & \textbf{K12} & \textbf{K15} & \textbf{Mid} & \textbf{ETH3D} & \textbf{Mean} \\
  \hline\hline
  Single Best
  & 4.13 & \underline{3.64} & 6.56 & 2.76 & 4.27
  & 3.48 & 3.35 & \underline{3.07} & 1.61 & \underline{2.88} \\
  Uniform Avg.
  & 3.29 & 4.08 & 6.61 & 2.27 & 4.06
  & 3.15 & 3.94 & 5.31 & 2.14 & 3.64 \\
  Task Arith.~\cite{ilharco2023editing}
  & 3.27 & 4.02 & 6.48 & 2.24 & 4.00
  & \textbf{2.21} & \underline{3.17} & 3.90 & 5.32 & 3.65 \\
  Model Soups~\cite{wortsman2022model}
  & \underline{3.18} & 3.85 & \underline{5.95} & \underline{2.22} & \underline{3.80}
  & 3.60 & 4.62 & 3.68 & \underline{1.19} & 3.27 \\
  TIES~\cite{yadav2023ties}
  & 3.42 & 4.06 & 7.17 & 2.85 & 4.38
  & 6.32 & 4.63 & 4.78 & 7.06 & 5.70 \\
  DARE~\cite{yu2024dare}
  & 3.39 & 4.15 & 7.01 & 2.36 & 4.23
  & 3.64 & 3.74 & 7.76 & 3.64 & 4.70 \\
  \hline
  \rowcolor{cyan!10}
  \textbf{StereoFactory (Ours)}
  & \textbf{3.15} & \textbf{3.29} & \textbf{5.26} & \textbf{1.51} & \textbf{3.30}
  & \underline{2.45} & \textbf{2.92} & \textbf{2.29} & \textbf{1.10} & \textbf{2.19} \\
  \thickhline
\end{tabular}}
\label{tab:merging_comparison}
\end{table*}

\begin{table}[t]
\centering
\caption{Training-free generalization results on four public datasets. The most commonly used metrics for each dataset are adopted, and lower is better. Rows marked with $^*$ use stronger multi-dataset or paper-specific training recipes and are listed as reference points. In each lower architecture-specific block, bold marks the better non-reference result between the original checkpoint and StereoFactory, independent of the starred reference rows. StereoFactory (Ours) performs post-hoc parameter merging after source checkpoints are available, without joint retraining or target-domain fine-tuning.}
\label{tab:zero_shot}
\setlength\tabcolsep{4pt}
\renewcommand\arraystretch{1.15}
\small
\resizebox{\columnwidth}{!}{
\begin{tabular}{l|ccccc}
\thickhline
\rowcolor[gray]{0.92}
 & \textbf{KITTI-12} & \textbf{KITTI-15} & \textbf{Middlebury} & \textbf{ETH3D} &  \\
\rowcolor[gray]{0.92}
\multirow{-2}{*}{\textbf{Methods}}
& \textbf{D1} & \textbf{D1} & \textbf{BP-2} & \textbf{BP-1} 
& \multirow{-2}{*}{\textbf{Mean}} \\
\hline\hline
CREStereo++~\cite{li2022crestereo} & 4.7 & 5.2 & 14.8 & 4.4 & 7.28\\
DSMNet~\cite{dsmnet2020}  & 6.2 & 6.5 & 13.8 & 6.2 & 8.18\\
Mask-CFNet~\cite{rao2023masked} & 4.8 & 5.8 & 13.7 & 5.7 & 7.50\\
HVT-RAFT~\cite{chang2023domain} & 3.7 & 5.2 & 10.4 & 3.0 & 5.58\\
RAFT-Stereo~\cite{lipson2021raft} & 4.7 & 5.5 & 12.6 & 3.3 & 6.52\\
Selective-IGEV~\cite{SelectiveStereo} & 4.5 & 5.6 & 9.2 & 5.7 & 6.25\\
IGEV~\cite{igev2023} & 5.2 & 5.7 & 8.8 & 4.0 & 5.92\\
Former-RAFT-DAM~\cite{zhang2025learning} & 3.9 & 5.1 & 8.1 & 3.3 & 5.10\\
IGEV++~\cite{xu2024igev++} & 5.1 & 5.9 & 7.8 & 4.1 & 5.72\\
NMRF~\cite{NMRFStereo} & 4.2 & 5.1 & 7.5 & 3.8 & 5.15\\
FoundationStereo (SceneFlow)~\cite{wen2025foundationstereo} & 3.2 & 4.9 & 5.5 & 1.8 & 3.85\\

FoundationStereo (paper)~\cite{wen2025foundationstereo}$^*$ & 2.3 & 2.8 & 1.1 & 0.5 & 1.67\\
Selective-IGEV~\cite{SelectiveStereo}$^*$ & 3.2 & 4.5 & 7.5 & 3.4 & 4.65\\
\hline\hline

NMRF-SwinT~\cite{NMRFStereo} & 8.67 & 7.46 & 16.36 & 23.46 & 13.99\\
\rowcolor{gray!10}
NMRF-SwinT~\cite{NMRFStereo}$^*$ & 3.48 & 3.88 & 7.03 & 1.83 & 4.06\\
\rowcolor{cyan!10}
\textbf{NMRF-SwinT~\cite{NMRFStereo}  (Ours)} 
& \textbf{3.15} & \textbf{3.29} & \textbf{5.26} & \textbf{1.51} & \textbf{3.30}\\

FoundationStereo~\cite{wen2025foundationstereo} & 3.2 & 4.9 & 5.5 & 1.8 & 3.85\\
\rowcolor{gray!10}
FoundationStereo~\cite{wen2025foundationstereo}$^*$ & 3.01 & 3.26 & 2.69 & 0.77 & 2.43\\
\rowcolor{cyan!10}
\textbf{FoundationStereo~\cite{wen2025foundationstereo} (Ours)} 
& \textbf{2.45} & \textbf{2.92} & \textbf{2.29} & \textbf{1.10} & \textbf{2.19}\\
\thickhline
\end{tabular}%
}
\vspace{-10pt}
\end{table}

% ============================================
% C. Comparison with Reference Methods
% ============================================
\subsection{Comparison with Strong Reference Methods}
\label{subsec:sota}
Table~\ref{tab:zero_shot} presents training-free generalization results on four
public benchmarks. The upper block reports published stereo baselines and
stronger reference configurations, while the lower block reports the two
backbones used in our merging study; when a backbone also appears in the reference
block, the lower block repeats the checkpoint corresponding to our merging setup
for clarity. Rows marked with $^*$ employ stronger
multi-dataset training recipes or paper-specific data combinations and are
included as reference points rather than directly comparable post-hoc merging
baselines. Nonetheless, StereoFactory establishes a strong post-hoc alternative
across both architectures. For NMRF, StereoFactory reaches an average error of 3.30,
improving over both the published reference row at 5.15 and the reproduced
multi-dataset checkpoint at 4.06, without joint retraining or target-domain
fine-tuning. For FoundationStereo, our method
attains an average error of 2.19, improving over the SceneFlow-only baseline
at 3.85 and the reproduced multi-dataset checkpoint at 2.43 in our setting.
The stronger FoundationStereo (paper) row uses a different training recipe and is
therefore reported for reference rather than as a directly comparable post-hoc
merging baseline.

Table~\ref{tab:merging_comparison} further compares merging strategies under
controlled conditions, where all methods operate on the identical set of 11
checkpoints. StereoFactory attains the lowest average error on every
architecture: 3.30 on NMRF versus 3.80 for Model Soups, and 2.19 on
FoundationStereo versus 2.88 for Single Best. The column-wise highlights in the table indicate per-metric
best/second-best values; consequently, some individual metrics are led by other
baselines, while the average columns consistently favor StereoFactory.
Uniform Averaging performs reasonably on NMRF with an average of 4.06, but still
lags behind StereoFactory by 0.76 average points. Parameter-level
interference-reduction methods exhibit inconsistent behavior: TIES and DARE do
not reliably improve over uniform averaging, and Model Soups remains competitive
on NMRF but is weaker than StereoFactory on both architectures. These results
indicate that StereoFactory is effective because Stage~1 reduces harmful
source-domain interference and Stage~2 exploits module-level specialization within
the selected pool, making it more reliable than merge-all or greedy alternatives.

\begin{figure*}[!t]
    \centering
    \scriptsize
    \setlength{\tabcolsep}{1pt}
    \renewcommand{\arraystretch}{1.0}
    \begin{tabular}{c c c c c c}
        & \textbf{RGB} & \textbf{NMRF-Mean} & \textbf{NMRF-Ours} & \textbf{FoundationS-Mean} & \textbf{FoundationS-Ours} \\

        \raisebox{8mm}{\rotatebox[origin=c]{90}{\textbf{KITTI12}}} &
        \includegraphics[width=0.18\textwidth,height=18mm]{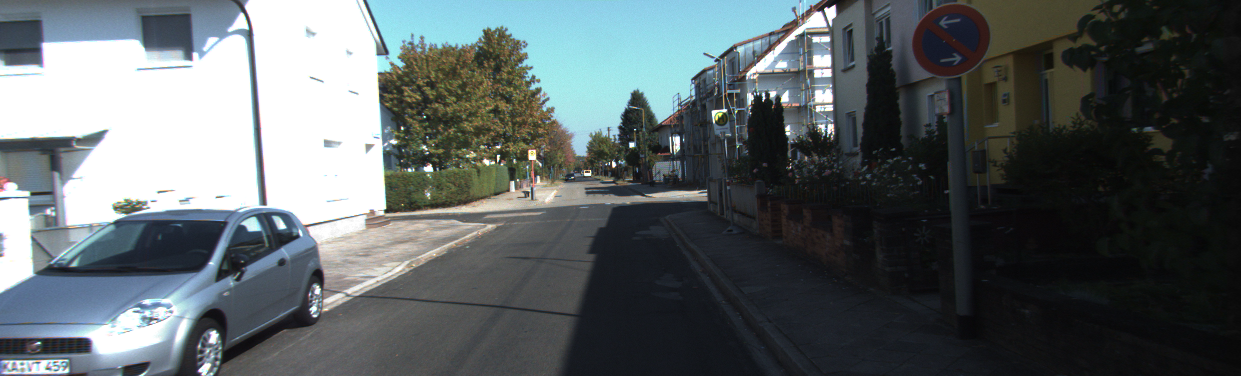} &
        \includegraphics[width=0.18\textwidth,height=18mm]{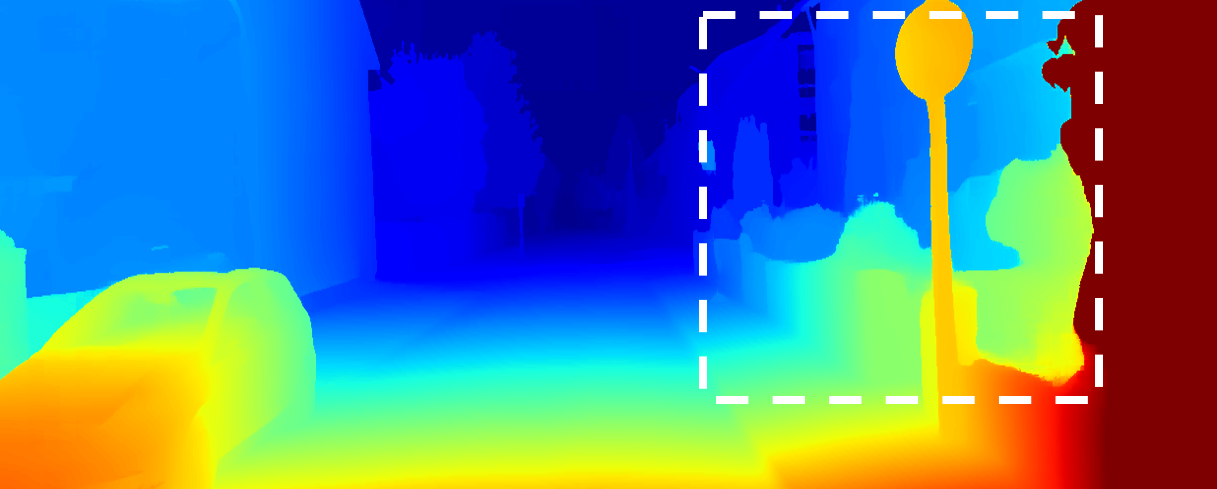} &
        \includegraphics[width=0.18\textwidth,height=18mm]{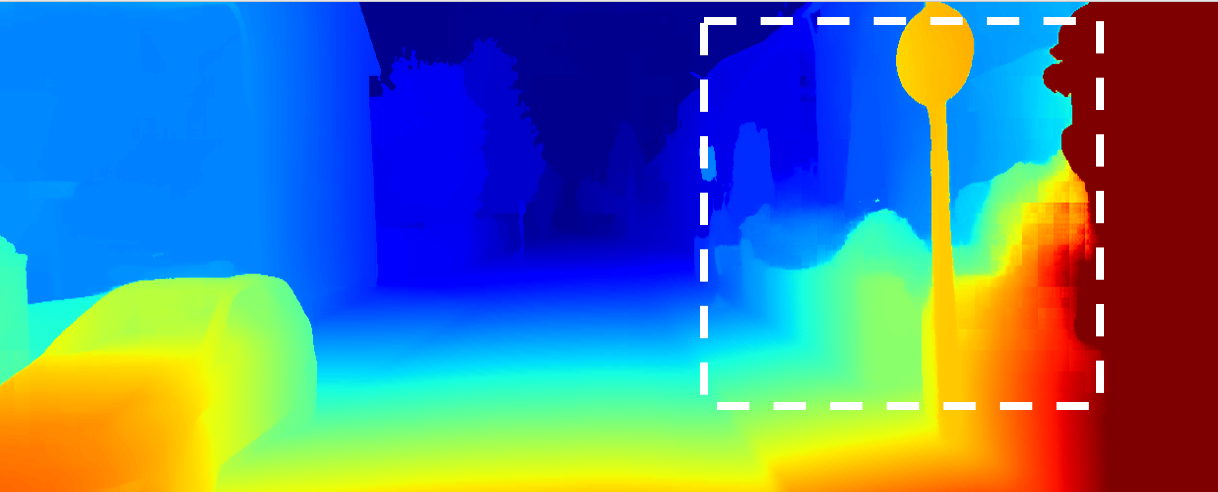} &
        \includegraphics[width=0.18\textwidth,height=18mm]{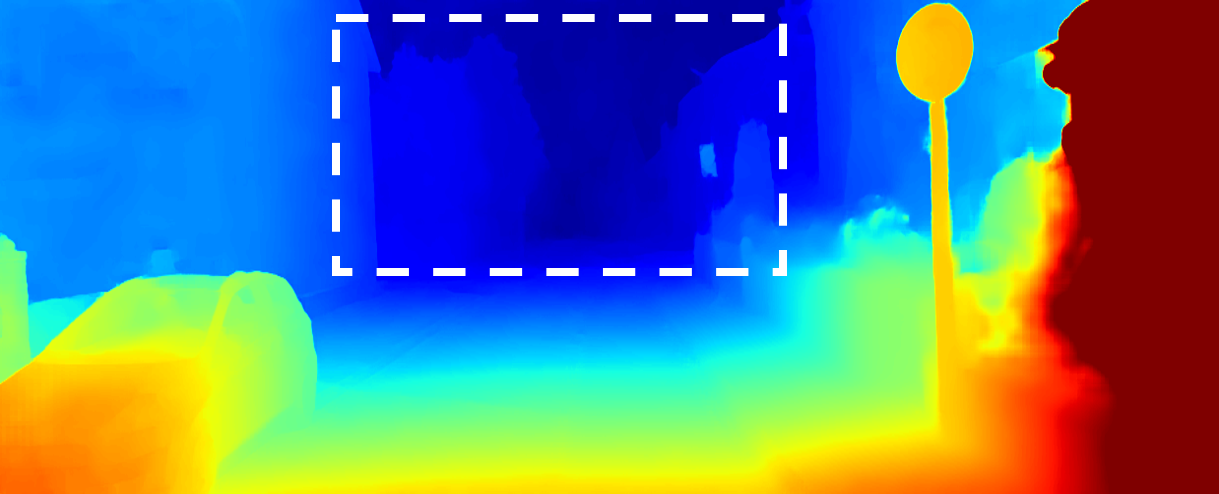} &
        \includegraphics[width=0.18\textwidth,height=18mm]{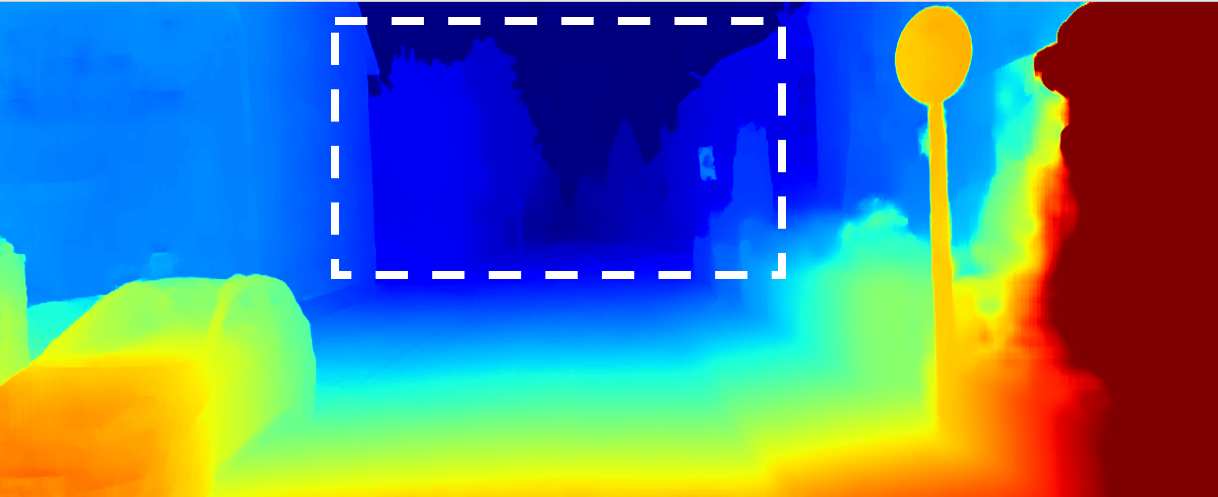} \\

        \raisebox{8mm}{\rotatebox[origin=c]{90}{\textbf{KITTI15}}} &
        \includegraphics[width=0.18\textwidth,height=18mm]{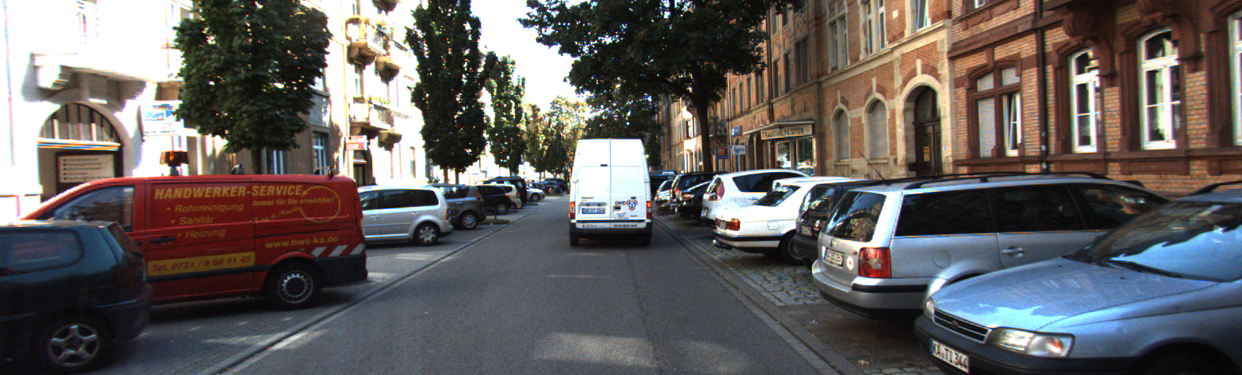} &
        \includegraphics[width=0.18\textwidth,height=18mm]{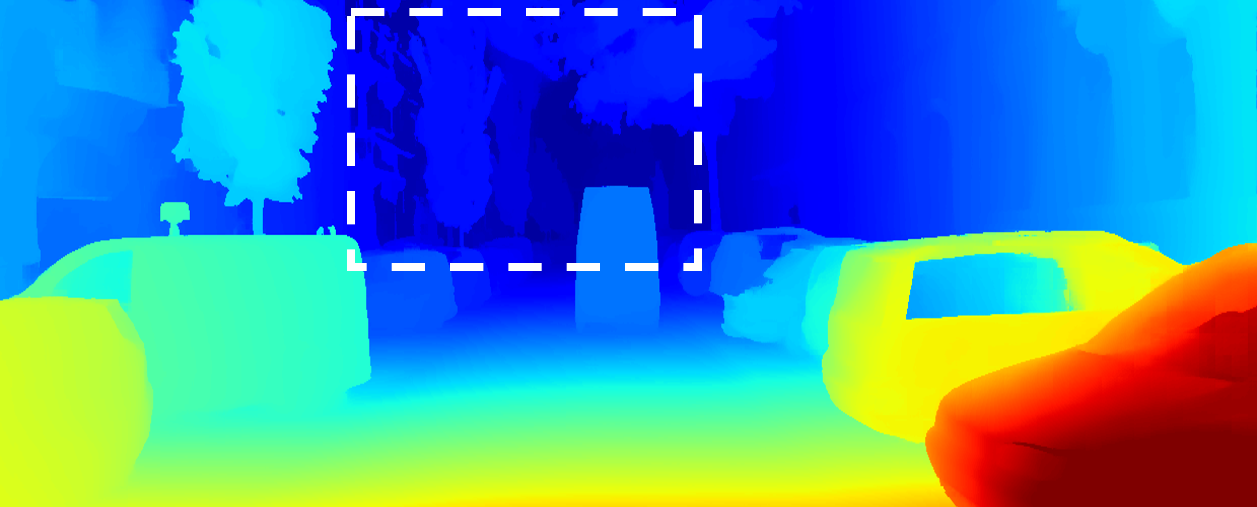} &
        \includegraphics[width=0.18\textwidth,height=18mm]{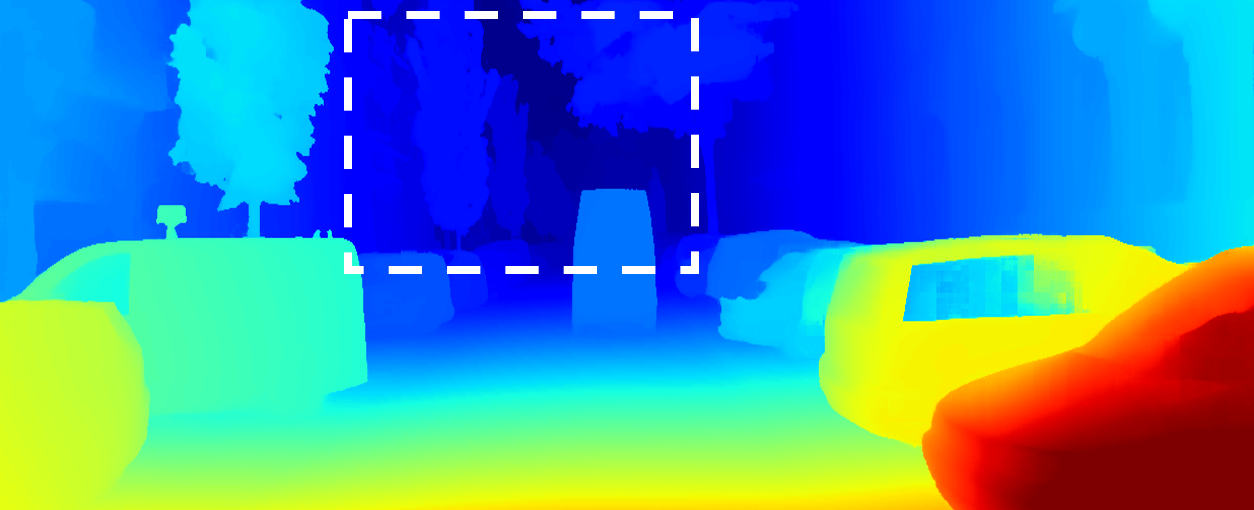} &
        \includegraphics[width=0.18\textwidth,height=18mm]{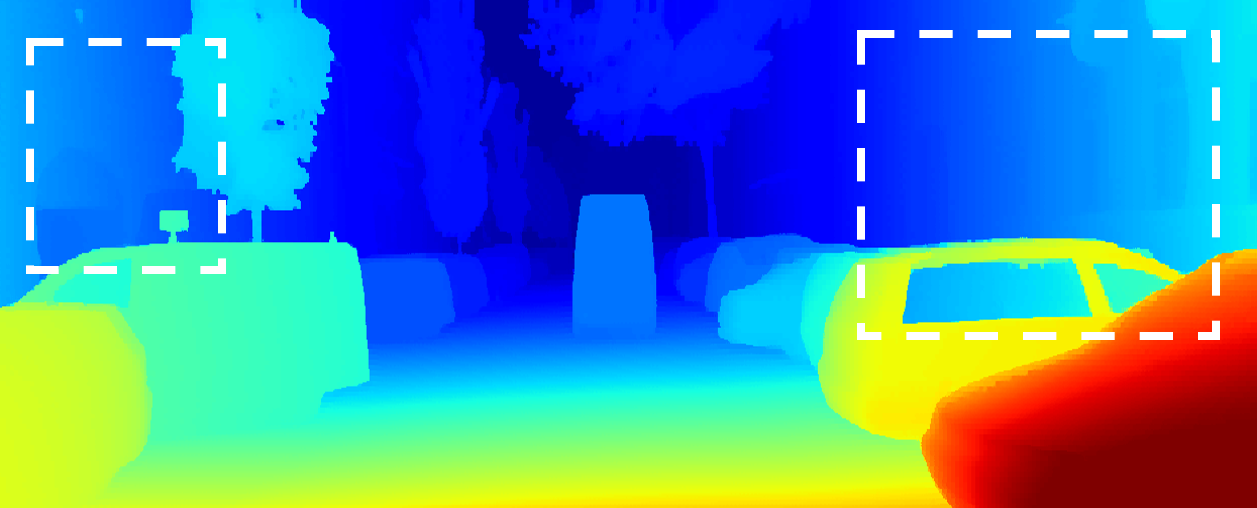} &
        \includegraphics[width=0.18\textwidth,height=18mm]{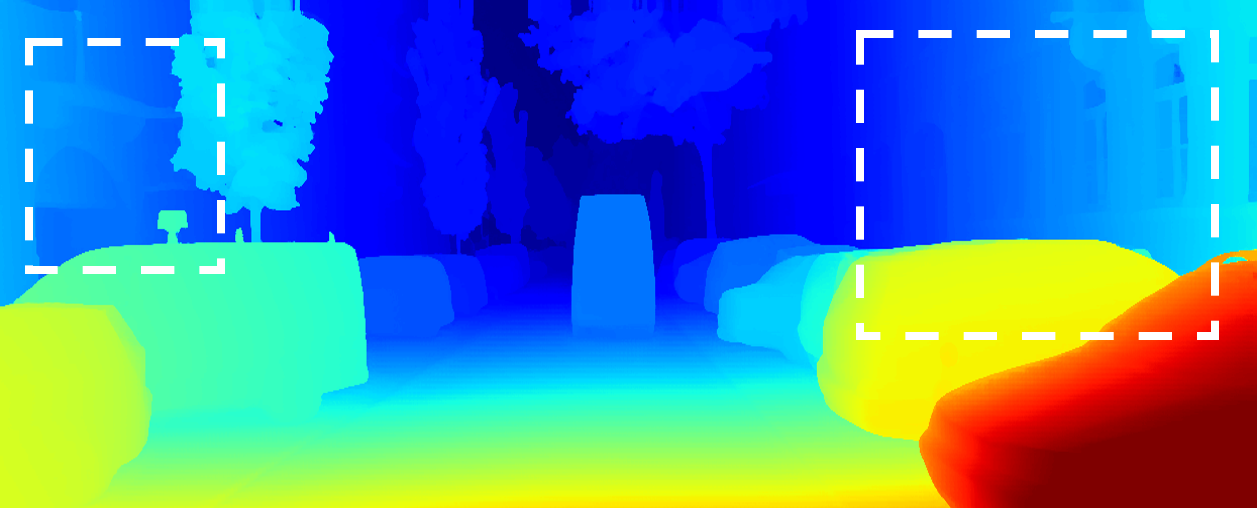} \\

        \raisebox{8mm}{\rotatebox[origin=c]{90}{\textbf{Middlebury}}} &
        \includegraphics[width=0.18\textwidth,height=18mm]{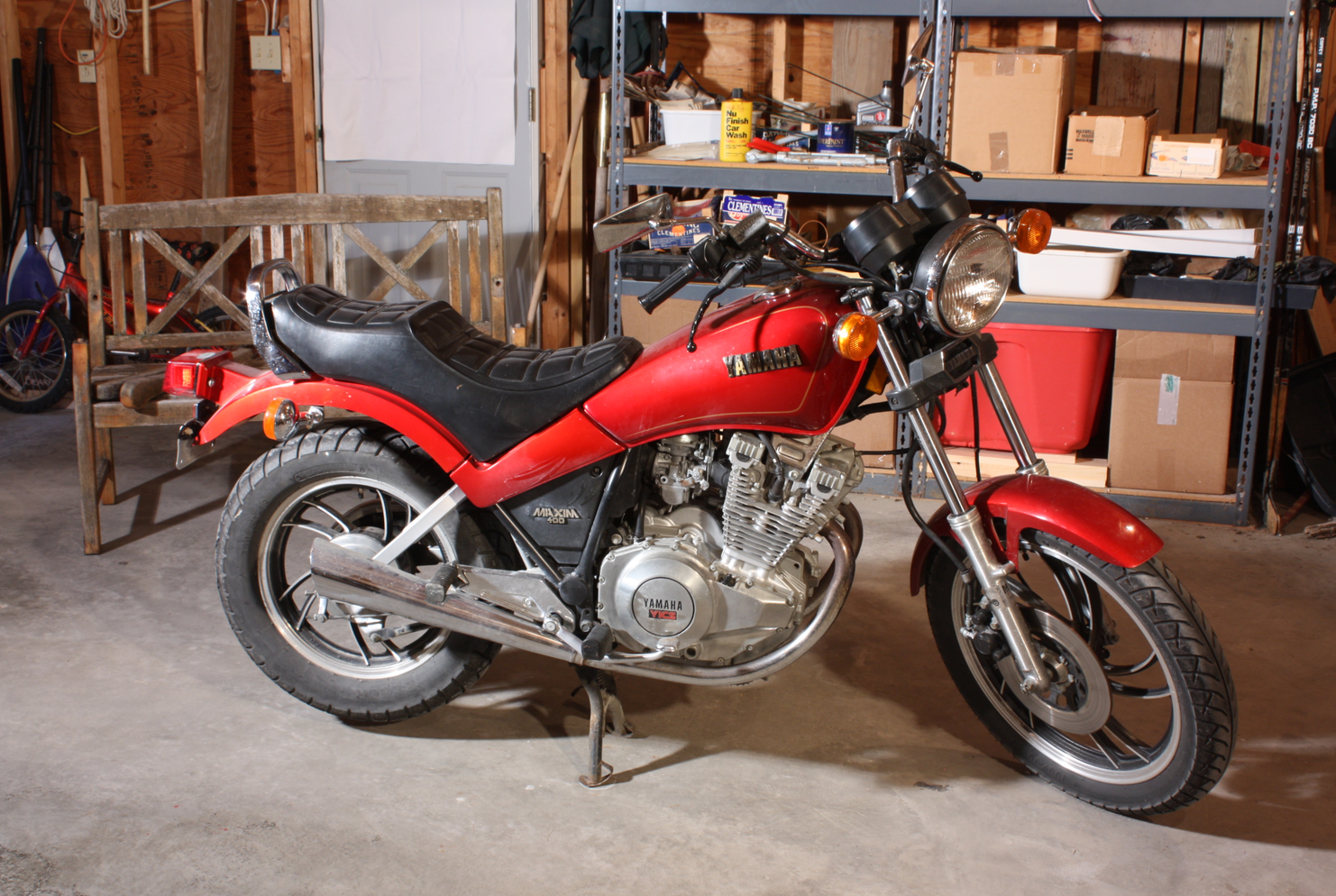} &
        \includegraphics[width=0.18\textwidth,height=18mm]{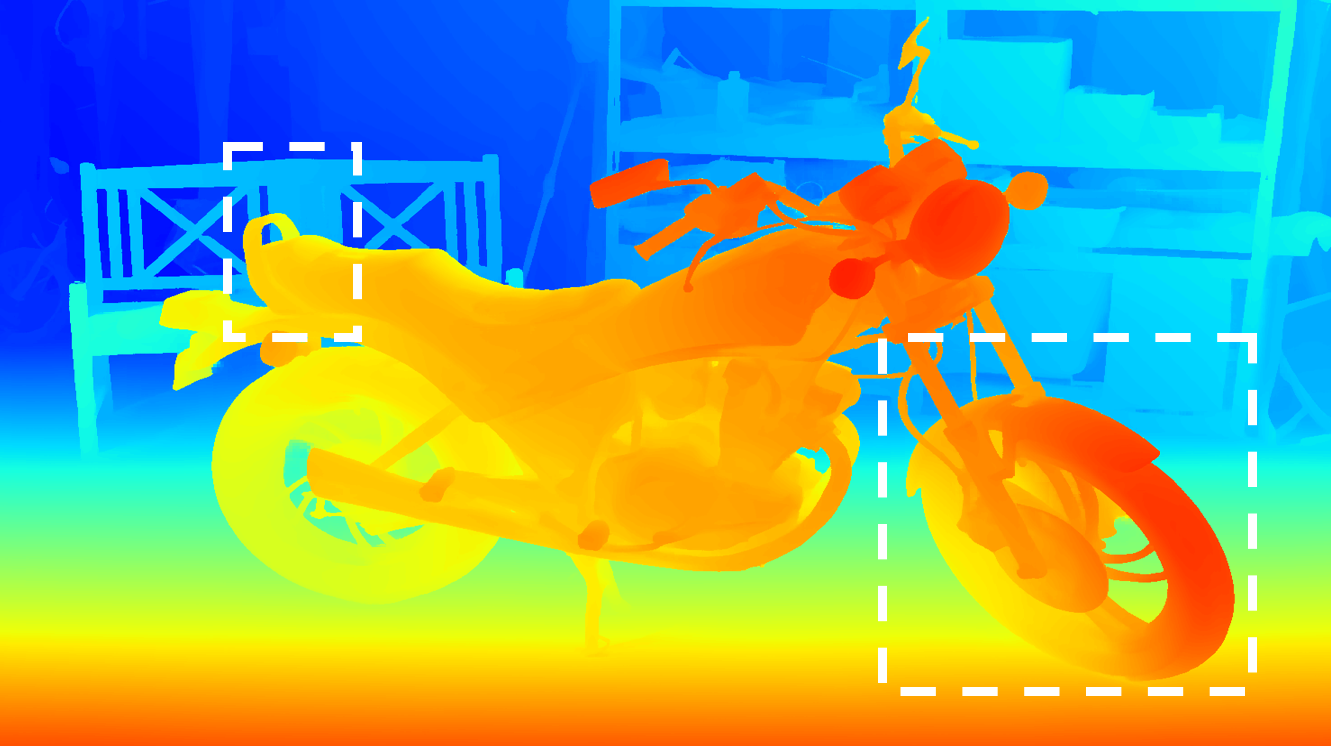} &
        \includegraphics[width=0.18\textwidth,height=18mm]{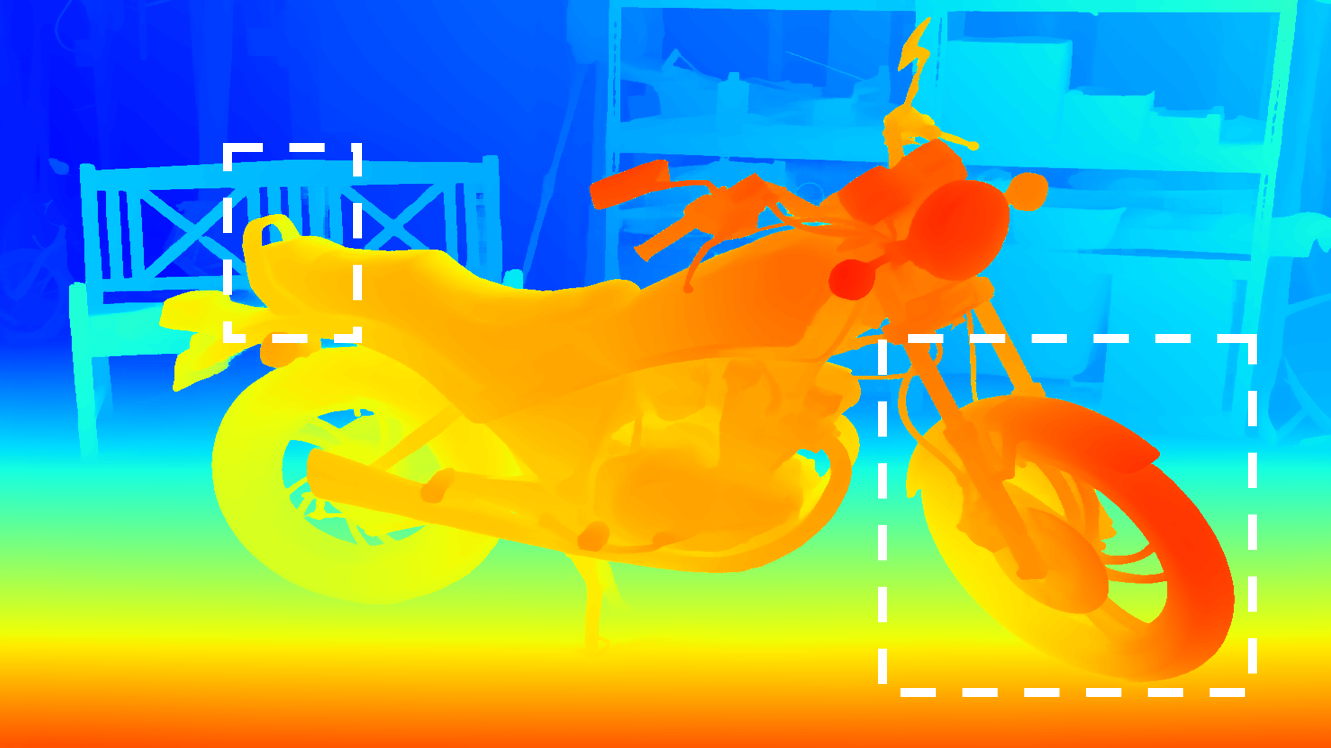} &
        \includegraphics[width=0.18\textwidth,height=18mm]{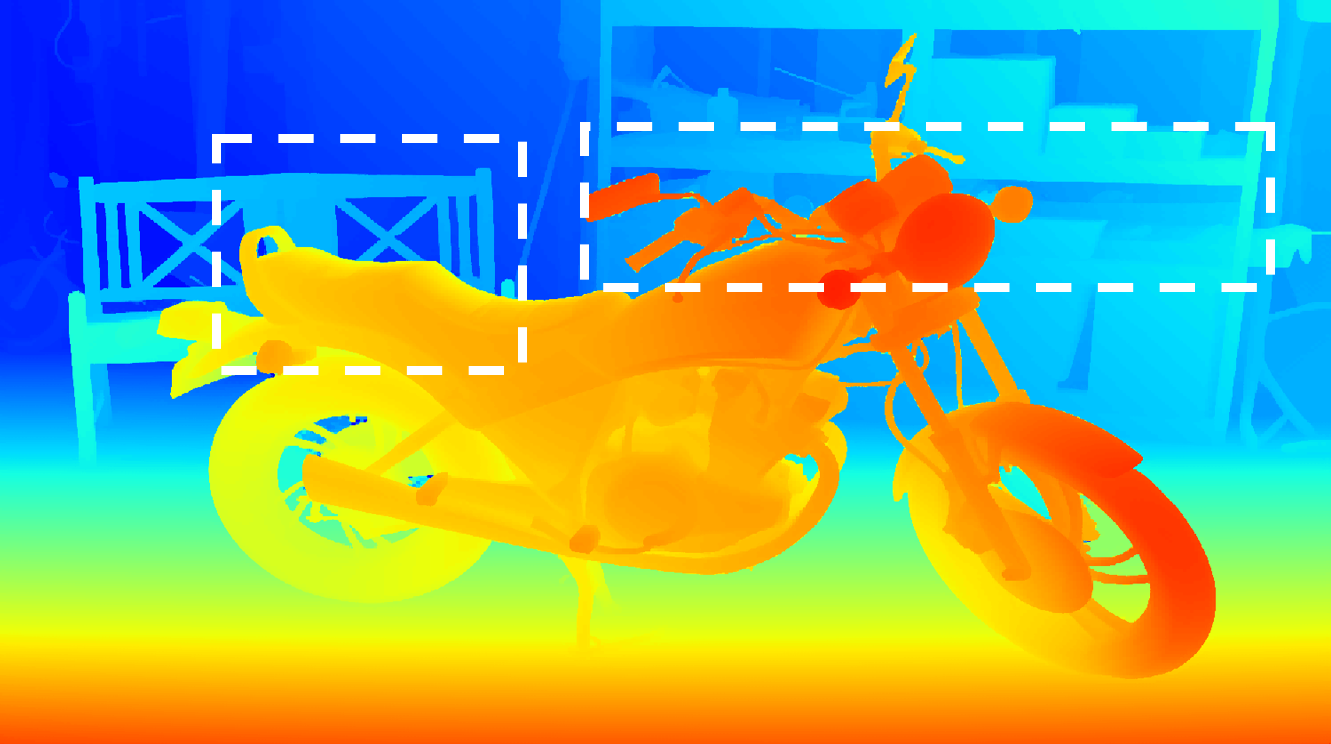} &
        \includegraphics[width=0.18\textwidth,height=18mm]{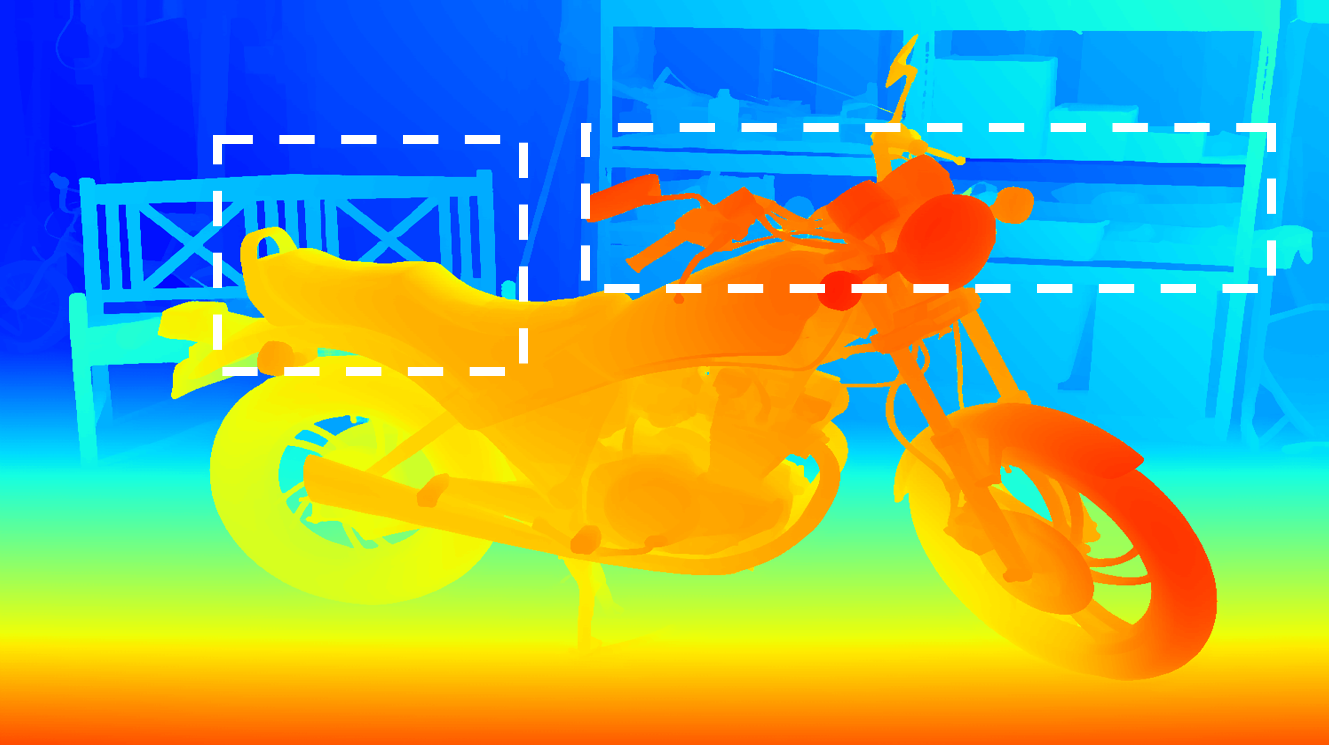} \\

        \raisebox{8mm}{\rotatebox[origin=c]{90}{\textbf{ETH3D}}} &
        \includegraphics[width=0.18\textwidth,height=18mm]{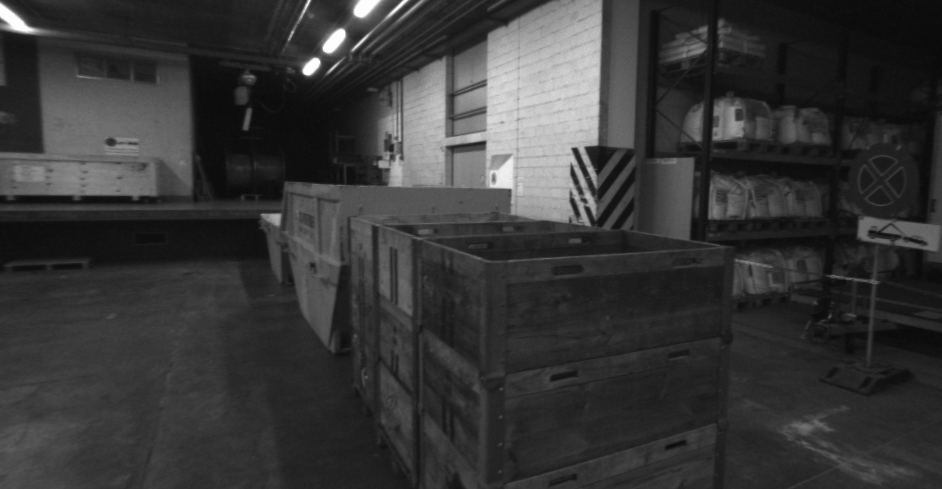} &
        \includegraphics[width=0.18\textwidth,height=18mm]{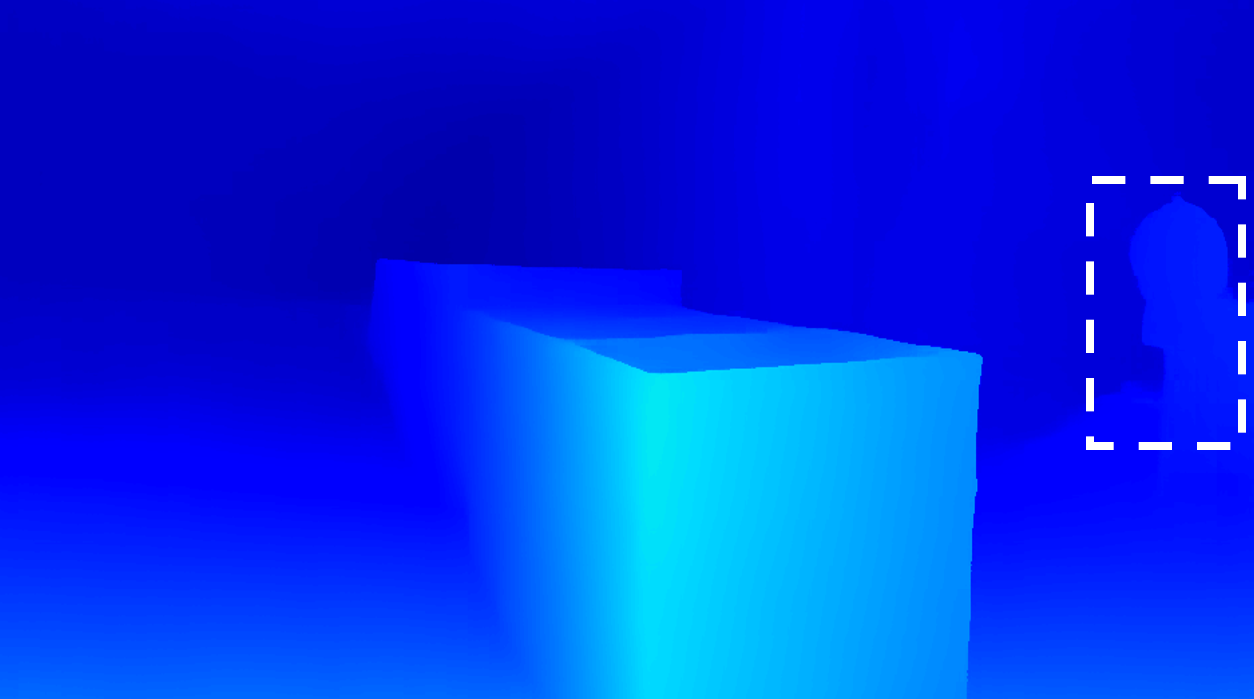} &
        \includegraphics[width=0.18\textwidth,height=18mm]{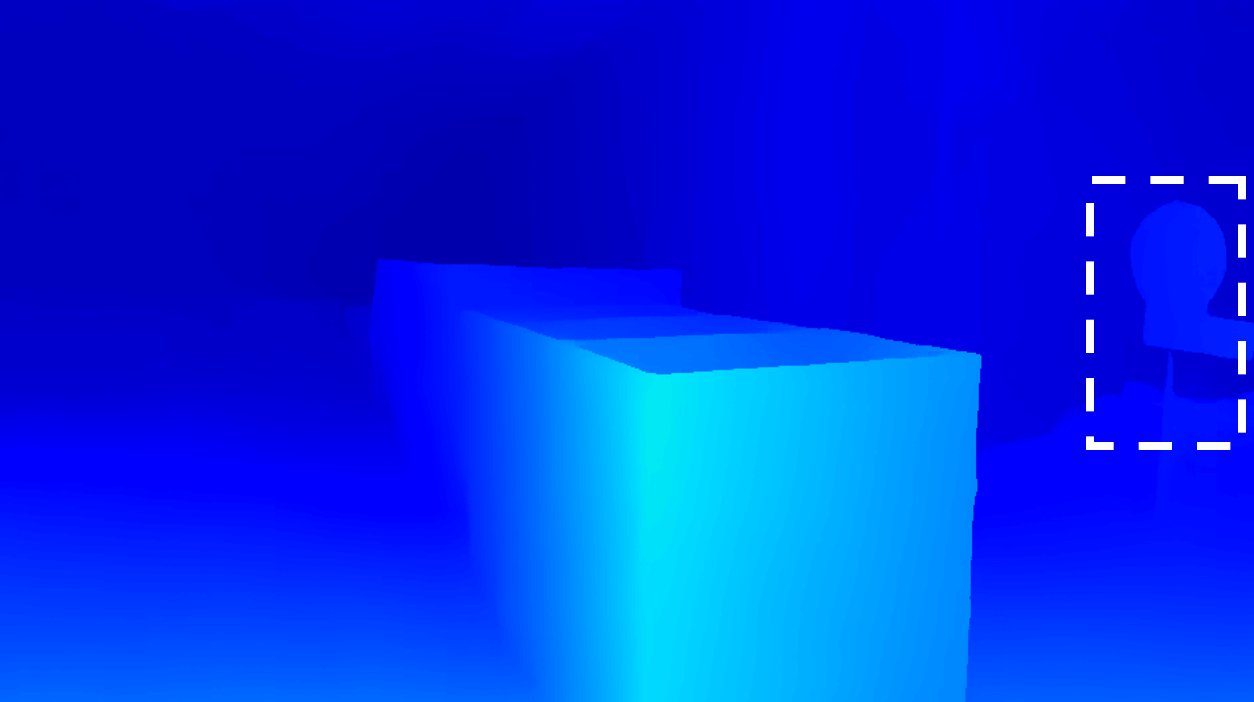} &
        \includegraphics[width=0.18\textwidth,height=18mm]{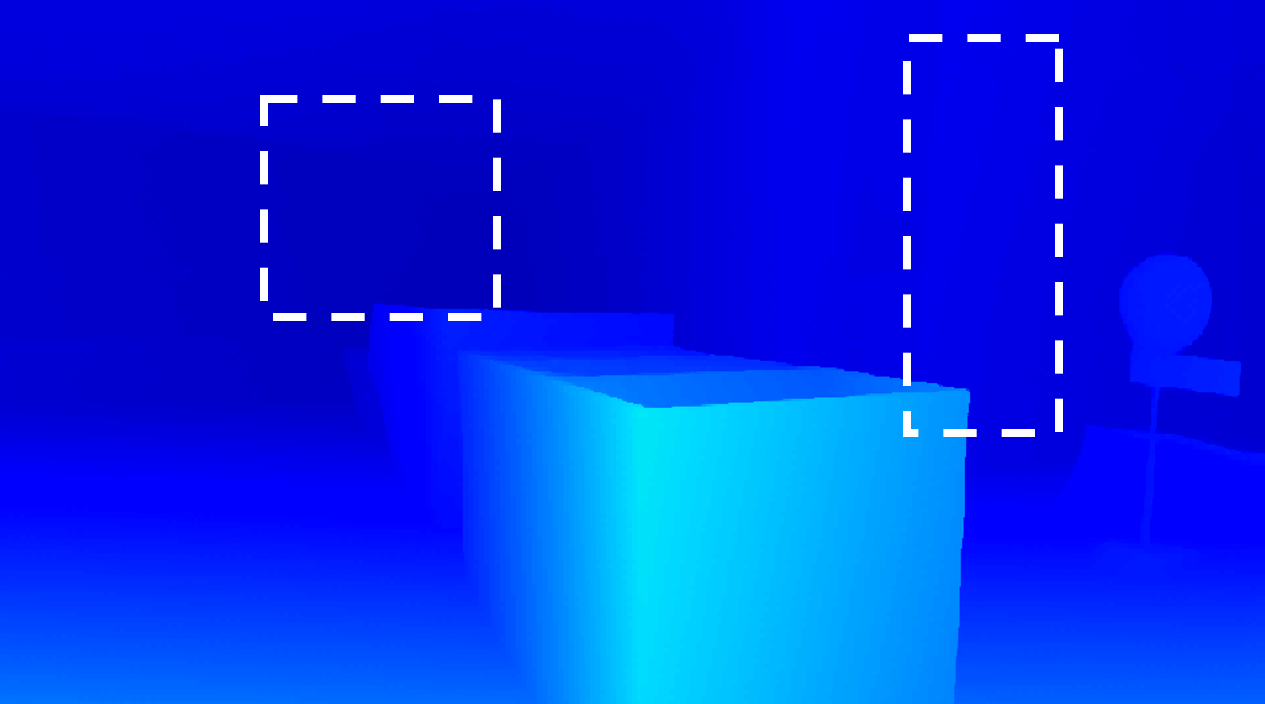} &
        \includegraphics[width=0.18\textwidth,height=18mm]{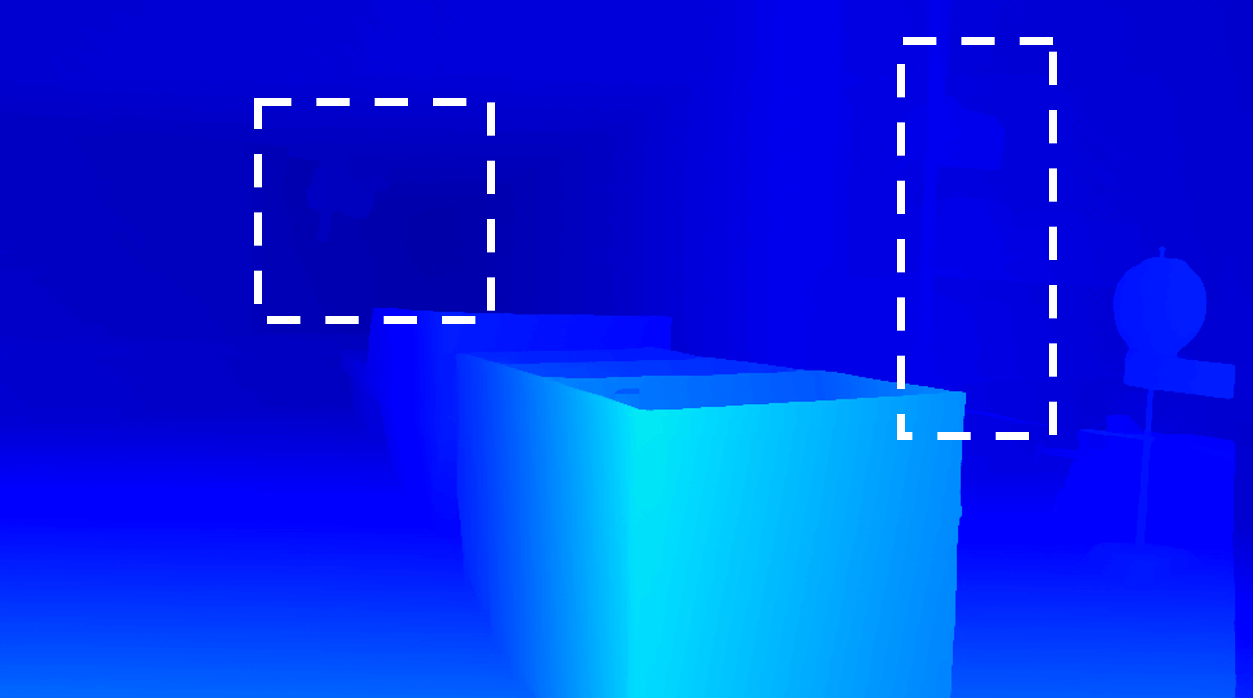} \\
    \end{tabular}

    \caption{\textbf{Qualitative comparison across NMRF-Stereo and
    FoundationStereo.} We compare uniform averaging (Mean) with StereoFactory
    (Ours) on representative samples from four target benchmarks. StereoFactory
    produces cleaner structures and fewer localized artifacts across different
    architectural families.}
    \label{fig:vis}
\end{figure*}

Fig.~\ref{fig:vis} provides qualitative examples on KITTI 2012, KITTI 2015,
Middlebury, and ETH3D. Compared with uniform averaging, StereoFactory preserves
cleaner disparity transitions and removes localized artifacts, visually consistent with
the average gains reported in Table~\ref{tab:merging_comparison}.

% ============================================
% D. Analysis of Individual Source Models
% ============================================
\subsection{Analysis of Individual Source Models}
\label{subsec:individual_analysis}

Tables~\ref{tab:individual_datasets} and~\ref{tab:benchmark_results} provide the
detailed per-checkpoint and dataset-combination results used in the following
analysis. Three principal findings emerge.

\textbf{Cross-dataset generalization varies drastically, with no
universally optimal source.}
The optimal training source varies as a function of the evaluation domain. On
NMRF, the checkpoint fine-tuned on the FoundationStereo training dataset gives
the lowest individual-checkpoint ETH3D error of 2.76, whereas VirtualKITTI2 leads
on KITTI\,12 at 3.96 yet degrades to 73.78 on ETH3D. FoundationStereo also shows source-specific outliers despite its
larger capacity: UnrealStereo4K reaches 59.23 on ETH3D, whereas the best sources
remain near 2.0. These harmful checkpoints explain why merge-all baselines can
underperform even when many individual checkpoints are useful.

\textbf{Sensitivity to dataset choice is architecture-dependent.}
NMRF exhibits large cross-dataset swings, particularly on ETH3D where source
errors range from 2.76 to 73.78. FoundationStereo, with its substantially larger
model capacity, demonstrates smaller but still meaningful fluctuations: KITTI\,12
errors range from 2.13 to 6.87 and Middlebury from 3.07 to 13.57. Thus, selective
merging benefits both a structured matching network and a foundation-style model,
although the magnitude of interference differs across backbones.

\begin{figure}[t]
  \centering
  \includegraphics[width=\columnwidth]{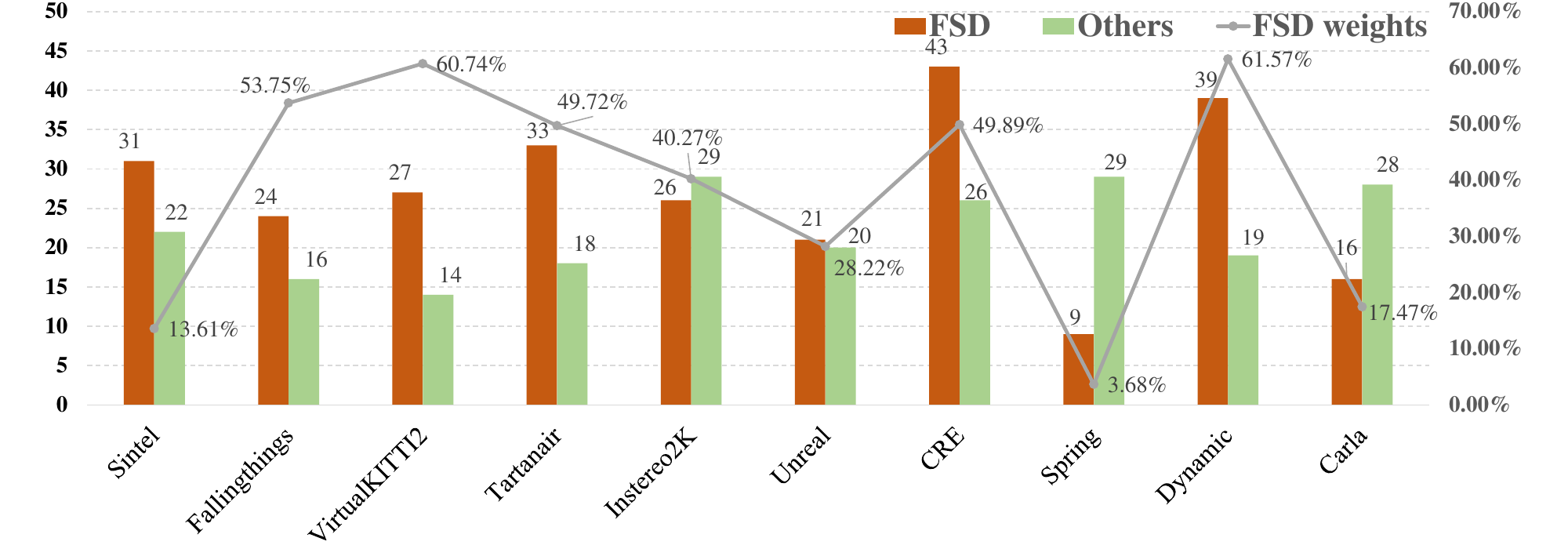}
  \vspace{-20pt}
  \caption{\textbf{Selection statistics of the FoundationStereo training dataset.}
  Across validation settings where the FoundationStereo training dataset is in
  the candidate pool, we compare its selection frequency and average fusion
  weight against the strongest competing source. The dataset is frequently
  selected with non-trivial weights, indicating that it provides a useful but not
  universally dominant source of transferable knowledge.}
  \label{fig:fsd_selection_statistics}
  \vspace{-20pt}
\end{figure}

\textbf{Role of the FoundationStereo Training Dataset in Merging.}
The detailed per-checkpoint results show that the checkpoint fine-tuned on the
FoundationStereo training dataset is often among the most transferable sources, especially
on ETH3D and Middlebury. This suggests that the representations acquired from
the FoundationStereo training dataset provide a useful anchor for multi-model merging,
although the final subset still depends on the architecture and validation
criterion. Fig.~\ref{fig:fsd_selection_statistics} further quantifies this trend:
the FoundationStereo training dataset is selected frequently and receives
substantial fusion weights in several validation settings, but it is not
universally dominant, reinforcing the need for adaptive subset search.

\begin{table*}[t]
\centering
\caption{\textbf{Performance of individual checkpoints and uniform averaging.}
Lower is better for all metrics.
\textbf{Bold} and \underline{underline} mark the best and second-best values
within each architecture-benchmark row.}
\setlength\tabcolsep{4pt}
\renewcommand\arraystretch{1.15}
\small
\resizebox{\textwidth}{!}{
\begin{tabular}{l|c|ccccccccccc|c}
  \thickhline
  \rowcolor[gray]{0.92}
  \textbf{Bench.}  & \textbf{SF}~\cite{sceneflow} & \textbf{Sintel}~\cite{butler2012naturalistic} & \textbf{FT}~\cite{tremblay2018falling}  & \textbf{VK2}~\cite{cabon2020virtual} & \textbf{Tar}~\cite{wang2020tartanair} & \textbf{I2K}~\cite{bao2020instereo2k} & \textbf{U4K}~\cite{tosi2021smdnets} & \textbf{CRE}~\cite{li2022crestereo} & \textbf{SP}~\cite{mehl2023spring} & \textbf{DY}~\cite{karaev2023dynamicstereo} & \textbf{Carla}~\cite{guo2025stereocarla} & \textbf{FS}~\cite{wen2025foundationstereo} & \textbf{Uniform Avg.} \\
  \hline\hline

  \multicolumn{14}{l}{\textbf{NMRF~\cite{NMRFStereo}}}\\\hline
  KITTI12~\cite{kitti2012}
  & 8.67 & 6.09 & 4.28 & \underline{3.96} & 4.16 & 13.33 & 8.68 & 8.01 & 6.59 & 11.84 & 4.11 & 4.13 & \textbf{3.29} \\

  KITTI15~\cite{menze2015kitti15}
  & 7.46 & 6.28 & 4.23 & \underline{4.00} & 4.71 & 15.21 & 6.90 & 6.18 & 6.23 & 15.36 & 4.87 & \textbf{3.64} & 4.08 \\

  Midd.~\cite{scharstein2014middlebury}
  & 16.36 & 19.28 & 13.17 & 22.23 & 13.95 & 11.75 & 44.98 & 13.73 & 16.04 & 12.84 & 9.12 & \textbf{6.56} & \underline{6.61} \\

  ETH3D~\cite{schops2017eth3d}
  & 23.46 & 6.18 & 27.93 & 73.78 & 5.25 & 11.23 & 64.51 & 5.75 & 6.96 & 5.32 & 3.17 & \underline{2.76} & \textbf{2.27} \\

  % \hline
  % \multicolumn{14}{l}{\textbf{LightStereo~\cite{guo2025lightstereo}}}\\\hline
  % Kitti12~\cite{kitti2012}
  % & 6.40 & 6.28 & \underline{4.19} & 4.42 & 4.43 & 13.86 & 7.88 & 16.91 & 8.52 & 8.93 & 5.09 & 5.02 & \textbf{3.95} \\

  % Kitti15~\cite{menze2015kitti15}
  % & 6.40 & 7.03 & 4.75 & \textbf{3.94} & 5.42 & 17.19 & 6.89 & 14.73 & 8.27 & 11.72 & 6.37 & 5.00 & \underline{4.52} \\

  % Midd~\cite{scharstein2014middlebury}
  % & 17.51 & 18.3 & 12.28 & 18.62 & 13.46 & 16.78 & 17.4 & 20.66 & 20.26 & 13.75 & 13.41 & \textbf{10.53} & \underline{12.21} \\

  % ETH3D~\cite{schops2017eth3d}
  % & 11.37 & 7.62 & 33.88 & 9.6 & \textbf{5.97} & 67.36 & 46.74 & \underline{6.14} & 10.98 & 21.72 & 7.47 & 7.72 & 18.66 \\

  \hline
  \multicolumn{14}{l}{\textbf{FoundationStereo~\cite{wen2025foundationstereo}}}\\\hline
  KITTI12~\cite{kitti2012}
  & 5.53 & 4.45 & 3.50 & \underline{3.21} & 3.88 & 4.27 & 6.87 & 5.02 & 5.73 & 6.02 & 3.48 & 3.48 & \textbf{3.15} \\

  KITTI15~\cite{menze2015kitti15}
  & 5.93 & 5.13 & 4.20 & \textbf{3.25} & 4.55 & 4.74 & 5.93 & 5.02 & 6.04 & 6.57 & 4.91 & \underline{3.35} & 3.94 \\

  Midd.~\cite{scharstein2014middlebury}
  & 7.69 & 11.17 & 5.54 & 11.82 & \underline{5.18} & 8.01 & 6.86 & 8.16 & 13.57 & 6.83 & 7.28 & \textbf{3.07} & 5.31 \\

  ETH3D~\cite{schops2017eth3d}
  & 2.13 & 2.13 & 26.19 & 3.93 & 2.54 & 4.22 & 59.23 & 4.54 & 4.68 & 5.93 & \underline{2.02} & \textbf{1.61} & 2.14 \\

  \thickhline
\end{tabular}}
\label{tab:individual_datasets}
\end{table*}

\begin{table*}[!t]
\centering
\caption{\textbf{Benchmark results with different dataset combinations.}
Lower is better for all metrics.
``Adapt'' denotes StereoFactory's adaptive search result. \textbf{Bold} and
\underline{underline} mark the best and second-best values within each
architecture-benchmark row.}
  \setlength\tabcolsep{4pt}
  \renewcommand\arraystretch{1.15}
\small
\resizebox{\textwidth}{!}{
\begin{tabular}{l|c|cccc|cccc|cccc|c}
  \thickhline
  \rowcolor[gray]{0.92}
  & 
  & \multicolumn{4}{c|}{\textbf{Feature Similarity}}
  & \multicolumn{4}{c|}{\textbf{RGB Similarity}}
  & \multicolumn{4}{c|}{\textbf{Disparity Similarity}}
  & \\
  \rowcolor[gray]{0.92}
  \multirow{-2}{*}{\textbf{Bench.}}
  & \multirow{-2}{*}{\textbf{Uniform Avg.}}
  & \textbf{K12} & \textbf{K15} & \textbf{Mid} & \textbf{ETH3D}
  & \textbf{K12} & \textbf{K15} & \textbf{Mid} & \textbf{ETH3D}
  & \textbf{K12} & \textbf{K15} & \textbf{Mid} & \textbf{ETH3D}
  & \multirow{-2}{*}{\textbf{Adapt}} \\
  \hline\hline

  \multicolumn{15}{l}{\textbf{NMRF~\cite{NMRFStereo}}}\\\hline
  KITTI12~\cite{kitti2012} & 3.29 & \underline{3} & \underline{3} & 3.01 & 3.01 & 3.31 & 3.32 & 3.32 & 3.31 & 3.02 & 3.12 & \textbf{2.98} & 3.6 & 3.15\\
  KITTI15~\cite{menze2015kitti15} & 4.08 & 3.67 & 3.67 & 3.68 & 3.68 & 4.04 & 4.05 & 4.04 & 4.06 & 3.71 & 3.83 & \underline{3.66} & 4.34 & \textbf{3.29} \\
  Midd.~\cite{scharstein2014middlebury} & 6.61 & 6.64 & 6.63 & 6.64 & 6.64 & 6.79 & 6.78 & 6.77 & 6.73 & 6.6 & 6.67 & \underline{6.5} & 7.15 & \textbf{5.26}\\
  ETH3D~\cite{schops2017eth3d} & 2.27 & 2.31 & 2.32 & 2.31 & 2.31 & 2.27 & 2.26 & \underline{2.24} & 2.27 & 2.3 & 2.3 & 2.29 & 2.35 & \textbf{1.51}\\

  % \hline
  % \multicolumn{15}{l}{\textbf{LightStereo~\cite{guo2025lightstereo}}}\\\hline
  % Kitti12~\cite{kitti2012} & 3.95 & \underline{3.93} & \underline{3.93} & 3.95 & 3.95 & 3.99 & 3.99 & 4 & 3.98 & 3.94 & 3.97 & \textbf{3.91} & 4.29 & 4.26\\
  % Kitti15~\cite{menze2015kitti15} & 4.52 & 4.54 & 4.54 & 4.53 & 4.55 & 4.47 & \textbf{4.45} & 4.49 & \underline{4.46} & 4.53 & 4.62 & 4.52 & 5.23 & 5.43\\
  % Midd~\cite{scharstein2014middlebury} & 12.21 & 12.29 & 12.32 & 12.25 & 12.15 & 12.24 & 12.19 & 12.38 & \textbf{7.96} & 12.07 & 12.02 & 12.06 & 13.19 & \underline{11.02} \\
  % ETH3D~\cite{schops2017eth3d} & 18.66 & 16.14 & 15.6 & 18.56 & 17.55 & 22.83 & 23.17 & 22.34 & 22.35 & 17.67 & 15.9 & 18.45 & \underline{6.44} & \textbf{4.23}\\

  \hline
  \multicolumn{15}{l}{\textbf{FoundationStereo~\cite{wen2025foundationstereo}}}\\\hline
  KITTI12~\cite{kitti2012} & 3.15 & 3.17 & 3.16 & 3.17 & 3.21 & \underline{3.06} & 3.09 & 3.07 & 3.13 & 3.18 & 3.20 & 3.20 & 3.67 & \textbf{2.45}\\
  KITTI15~\cite{menze2015kitti15} & 3.94 & 3.90 & \underline{3.89} & 4.00 & 4.00 & 4.11 & 4.10 & 4.13 & 4.08 & 3.94 & 3.97 & 3.92 & 4.22 & \textbf{2.92}\\
  Midd.~\cite{scharstein2014middlebury} & 5.31 & 5.39 & 5.41 & 5.30 & 5.43 & 4.76 & 4.78 & \underline{4.66} & 5.03 & 5.31 & 5.30 & 5.29 & 6.33 & \textbf{2.29} \\
  ETH3D~\cite{schops2017eth3d} & 2.14 & 2.14 & 1.90 & 1.88 & 2.25 & 1.73 & 1.83 & \underline{1.67} & 2.08 & 2.19 & 2.15 & 2.27 & 3.00 & \textbf{1.10}\\
  \thickhline
\end{tabular}}
\label{tab:benchmark_results}
\end{table*}

% ============================================
% E. Dataset Combination Strategies
% ============================================
\subsection{Dataset Combination Strategies}
\label{subsec:combination}

Table~\ref{tab:benchmark_results} compares our adaptive search with three
heuristic subset strategies based on backbone feature similarity, raw RGB
similarity, and warped-disparity similarity. These target-specific heuristics can
occasionally win on individual benchmark--architecture pairs. For example,
disparity similarity gives the lowest NMRF KITTI 2012 score, while RGB
similarity remains competitive for several FoundationStereo settings. However,
the winning heuristic changes across benchmarks and architectures, and the same
rule can degrade on other targets. StereoFactory instead searches the subset and
fusion weights directly under the validation objective, yielding the best result
on seven of the eight architecture--benchmark pairs in
Table~\ref{tab:benchmark_results}. This pattern is consistent with the strongest
four-benchmark averages reported in Table~\ref{tab:merging_comparison} and
confirms that static similarity cues are useful diagnostics but cannot substitute
for search-based optimization when source-domain interference is target- and
architecture-dependent.

\begin{table}[t]
\centering
\caption{\textbf{Fitness-metric ablation.} We compare different validation
metrics for the evolutionary search on NMRF and FoundationStereo.}
\setlength\tabcolsep{4pt}
\renewcommand\arraystretch{1.15}
\small
\resizebox{\columnwidth}{!}{
\begin{tabular}{l|cccc|c}
  \thickhline
  \rowcolor[gray]{0.92}
  \textbf{Method} & \textbf{K12} & \textbf{K15} & \textbf{Mid} & \textbf{ETH3D} & \textbf{Mean} \\
  \hline\hline
  \multicolumn{6}{l}{\textit{NMRF~\cite{NMRFStereo}}} \\
  \hline
  Middlebury
    & 3.04 & 3.24 & 4.92 & 1.65 & 3.21 \\
  ETH3D
    & 3.15 & 3.29 & 5.26 & 1.51 & 3.30 \\
  \hline\hline
  \multicolumn{6}{l}{\textit{FoundationStereo~\cite{wen2025foundationstereo}}} \\
  \hline
  Middlebury
    & 2.80 & 3.06 & 2.02 & 0.91 & 2.20 \\
  ETH3D
    & 2.45 & 2.92 & 2.29 & 1.10 & 2.19 \\
  \thickhline
\end{tabular}}
\label{tab:ablation_1}
\end{table}

% ============================================
% F. Ablation Studies
% ============================================
\subsection{Ablation Studies}
\label{subsec:ablation}

\textbf{Sensitivity to Fitness Metric.}
Table~\ref{tab:ablation_1} examines the sensitivity of evolutionary search to the
choice of fitness metric. We execute the search separately with Middlebury Bad-2.0
and ETH3D Bad-1.0 as the fitness function, and evaluate the resulting merged models
on all four benchmarks. Across the evaluated architectures, the two fitness metrics
yield nearly identical outcomes. On NMRF, the Middlebury-guided search produces an
average error of 3.21 versus 3.30 for the ETH3D-guided search. On FoundationStereo,
the corresponding averages are 2.20 and 2.19, respectively. 
These results demonstrate that the evolutionary
search discovers dataset subsets exhibiting cross-metric robustness: distinct fitness
signals guide the search toward similar subset compositions rather than overfitting
to a particular validation criterion. In practice, this implies that the search is
not overly sensitive to the validation metric in the tested settings.

Table~\ref{tab:validation_subset_robustness} further exposes the actual subsets
found under different validation sets on NMRF. All three searches retain a highly
overlapping core of Sintel, TartanAir, Carla, and FoundationStereo, and their
four-benchmark averages remain within 0.01 of each other. This supports the view
that the evolutionary stage is primarily identifying reusable dataset
complementarity rather than exploiting a validation-specific artifact.

\begin{figure}[t]
  \centering
  \includegraphics[width=\columnwidth]{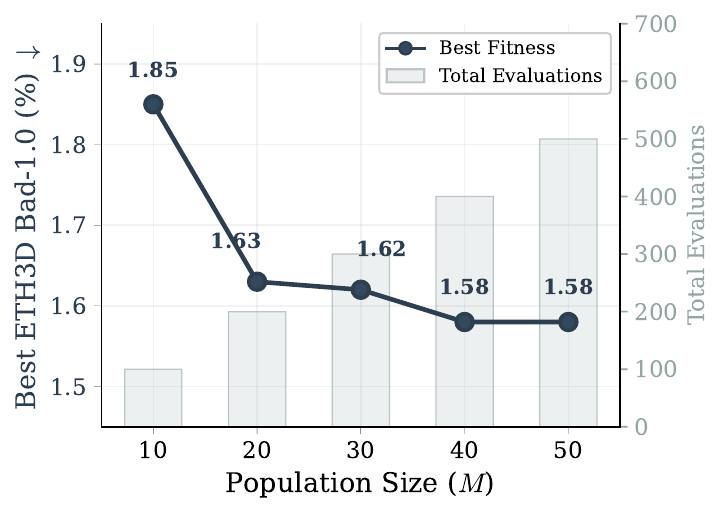}
  \vspace{-20pt}
  \caption{Population-size ablation for Stage~1 evolutionary search. Increasing
  the population size improves the best validation fitness until the search
  saturates around $M{=}40$, while larger populations mainly increase the total
  number of fitness evaluations. We use $M{=}50$ in the main experiments for a
  stable high-quality search.}
  \label{fig:pop_size_ablation}
\end{figure}

We also vary the Stage~1 population size in Fig.~\ref{fig:pop_size_ablation}.
The validation fitness improves rapidly when the population increases from a very
small search budget, then saturates around $M{=}40$, indicating that the
default setting provides stable search quality without unnecessary evaluations.

\begin{table}[t]
  \centering
  \caption{
    Validation-set robustness of evolutionary search on NMRF. Different
    validation sets discover highly overlapping subsets and produce nearly
    identical four-benchmark averages.
  }
  \label{tab:validation_subset_robustness}
  \setlength\tabcolsep{4pt}
  \renewcommand\arraystretch{1.15}
  \small
  \resizebox{\columnwidth}{!}{%
  \begin{tabular}{l|l|cccc|c}
    \thickhline
    \rowcolor[gray]{0.92}
    \textbf{Val Set} & \textbf{Subset} 
    & \textbf{K12} & \textbf{K15} & \textbf{Mid.} & \textbf{ETH3D} 
    & \textbf{Mean} \\
    \hline\hline
    ETH3D & \{0,3,7,9,10\} 
    & 3.15 & 3.29 & 5.26 & 1.51 & 3.30 \\
    
    Middlebury & \{0,2,3,9,10\} 
    & 3.18 & 3.31 & 5.18 & 1.58 & 3.31 \\
    
    KITTI~15 & \{0,3,7,9,10\} 
    & 3.12 & 3.24 & 5.31 & 1.54 & 3.30 \\
    \thickhline
  \end{tabular}}

  \vspace{1mm}
  \footnotesize
  Dataset indices: 0=Sintel, 2=VirtualKITTI2, 3=TartanAir, 7=Spring,
  9=Carla, 10=FS-train.
\end{table}

\textbf{Reusable Merging Across Architectures.}
Table~\ref{tab:ablation_2} examines whether the subset discovered on one
architecture transfers to another without re-executing the search. This is the
configuration we adopt by default for FoundationStereo, where the high retraining
cost makes subset reuse most valuable; we contrast it here with an
architecture-specific search to quantify any loss from transfer. The gap is
marginal in both directions. On NMRF, reusing the FoundationStereo-selected subset
attains an average error of 3.28 versus 3.30 for the architecture-specific search;
on FoundationStereo, reusing the NMRF-selected subset attains 2.19 versus 2.21 for
the architecture-specific search. Because the difference is within a few
hundredths and the ordering against all baselines is unchanged under either
choice, subset reuse incurs minimal degradation. These findings suggest that
reusable dataset complementarity, rather than purely architecture-specific
factors, drives the selected subsets, enabling a single search to serve multiple
deployment targets when the computational budget is constrained.

% ============================================================
% Stage 1 vs Stage 1+2 Ablation
% ============================================================

\begin{table}[t]
  \centering
  \caption{
    Ablation of the two-stage pipeline on NMRF. Stage~1 performs
    equal-weight subset selection via genetic algorithm; Stage~2
    applies the tied-routing hierarchical variant of our adaptive
    routing operator via CMA-ES. Stage~2 improves all four
    benchmarks.
  }
  \label{tab:s1_vs_s2}
  \setlength\tabcolsep{4pt}
  \renewcommand\arraystretch{1.15}
  \small
  \resizebox{\columnwidth}{!}{
  \begin{tabular}{l|cccc|c}
    \thickhline
    \rowcolor[gray]{0.92}
    \textbf{Method} & \textbf{K12} & \textbf{K15} & \textbf{Mid.} & \textbf{ETH3D} & \textbf{Mean} \\
    \hline\hline
    Stage 1 only   & 3.20 & 3.96 & 5.85 & 1.73 & 3.69 \\
    Stage 1 + 2    & 3.15 & 3.29 & 5.26 & 1.51 & 3.30 \\
    \thickhline
  \end{tabular}}
\end{table}

\textbf{Stage~1 vs.\ Stage~1+2.}
Table~\ref{tab:s1_vs_s2} isolates the contribution of Stage~2 adaptive routing
on NMRF. Stage~1 employs equal-weight averaging to identify the optimal dataset
subset, yielding an average error of 3.69. Stage~2 then optimizes the
tied-routing hierarchical variant: five selected-dataset weights (via softmax
parameterization) and four module-level scaling factors (via sigmoid
parameterization), using CMA-ES with ETH3D Bad-1.0 as the fitness criterion.
The adaptive routing stage reduces the average error from 3.69 to 3.30, with
consistent improvements across all four benchmarks: ETH3D decreases from 1.73
to 1.51, Middlebury from 5.85 to 5.26, KITTI\,12 from 3.20 to 3.15, and
KITTI\,15 from 3.96 to 3.29. Notably, all four benchmarks improve despite the
search being conducted solely on ETH3D, suggesting that adaptive routing
transfers beyond the validation metric. The learned module-level
scaling factors reveal interpretable specialization: the inference head
receives a scaling factor of 1.25, indicating that downstream disparity
refinement benefits from stronger task-specific adaptation, while the cost
volume module is scaled to 0.60, suggesting that matching cost computation
relies more heavily on the pretrained representations. These results confirm
that architecture-adaptive routing yields complementary gains beyond subset selection, thereby validating the efficacy of the proposed coarse-to-fine pipeline.

\begin{table}[t]
\centering
\caption{\textbf{Reusable-subset ablation.} We test whether the discovered subset
can transfer across architectures on NMRF and FoundationStereo.}
\setlength\tabcolsep{4pt}
\renewcommand\arraystretch{1.15}
\small
\resizebox{\columnwidth}{!}{
\begin{tabular}{l|cccc|c}
  \thickhline
  \rowcolor[gray]{0.92}
  \textbf{Method} & \textbf{K12} & \textbf{K15} & \textbf{Mid} & \textbf{ETH3D} & \textbf{Mean} \\
  \hline\hline
  \multicolumn{6}{l}{\textit{NMRF~\cite{NMRFStereo}}} \\
  \hline
  Non-reusable
    & 3.15 & 3.29 & 5.26 & 1.51 & 3.30 \\
  Reusable
    & 3.07 & 3.31 & 5.13 & 1.60 & 3.28 \\
  \hline\hline
  \multicolumn{6}{l}{\textit{FoundationStereo~\cite{wen2025foundationstereo}}} \\
  \hline
  Non-reusable
    & 2.85	& 3.23	& 1.91	& 0.85 &	2.21 \\
  Reusable
    & 2.45 & 2.92 & 2.29 & 1.10 & 2.19 \\
  \thickhline
\end{tabular}}
\label{tab:ablation_2}
\end{table}

\section{Conclusion}
\label{sec:conclusion}
This paper presents a coarse-to-fine evolutionary framework for adaptive model
merging in stereo matching. By decomposing the merging problem into discrete
subset selection via genetic algorithm and continuous architecture-adaptive
routing via CMA-ES, StereoFactory addresses the
what-to-merge and how-to-merge questions at progressively finer granularity,
without retraining after source checkpoints are available. Experiments
across two architectures and four benchmarks
demonstrate substantial average improvements over uniform merging and widely used
interference-reduction baselines, while adding only 2.7--3.7\% post-hoc search
overhead relative to joint retraining. Our analysis reveals that
knowledge contributions are inherently module-specific, with different functional
modules preferring distinct knowledge sources, and that selected subsets can
transfer across architectures with minimal degradation,
suggesting dataset complementarity is not purely architecture-specific.
These properties make the framework useful for incremental stereo model
maintenance: when new source data become available, one can train a
source-specific checkpoint and update the merged model through post-hoc search
rather than rerunning full multi-dataset training.
StereoFactory currently assumes weight-space compatibility, i.e., the checkpoints
being merged share the same architecture and base initialization, while the search
operator itself is reused across backbones. Future work includes automating the
block partitioning, combining adaptive selection with interference reduction,
and extending the framework to other tasks.

%%%%%%%%% REFERENCES

{\small
% \IEEEtriggeratref{73}
\bibliographystyle{ieee}
\bibliography{main}
}

% Put this in the preamble
\newcommand{\bioimg}[2][]{%
  \includegraphics[width=1in,height=1.25in,clip,#1]{#2}%
}

\vspace{-30pt}
\begin{IEEEbiography}
[{\bioimg{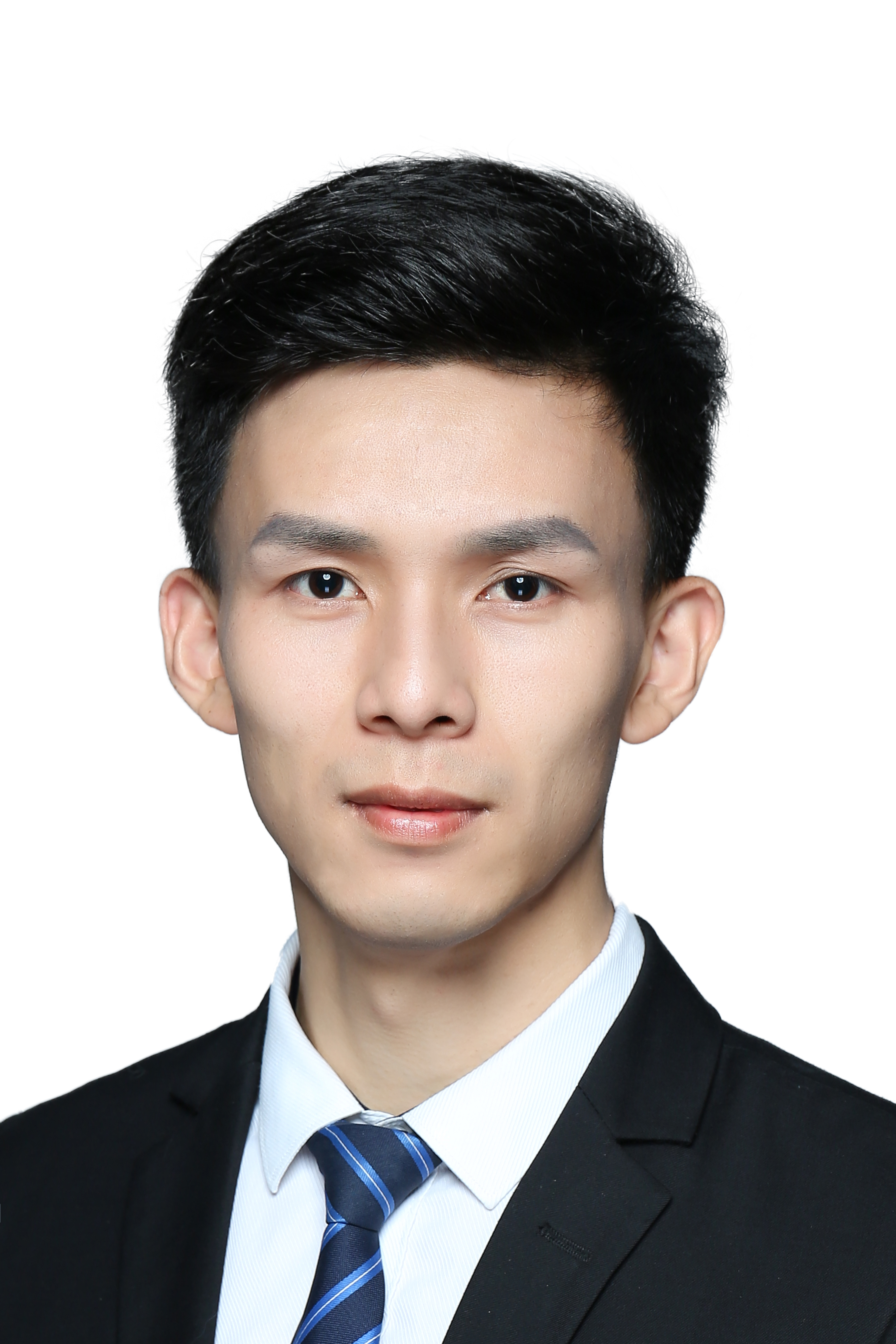}}]
{Xianda Guo} is currently a Ph.D. student at the School of Computer Science, Wuhan University. Before that, he received a master's degree from the Dalian University of Technology in 2019. His work is mainly in computer vision, such as gait recognition, depth estimation, and stereo-matching. He has co-authored publications on CVPR, ICCV, and ECCV. He is the creator of the widely used open-source platform OpenStereo.
\end{IEEEbiography}

\vspace{-40pt}
\begin{IEEEbiography}
[{\bioimg{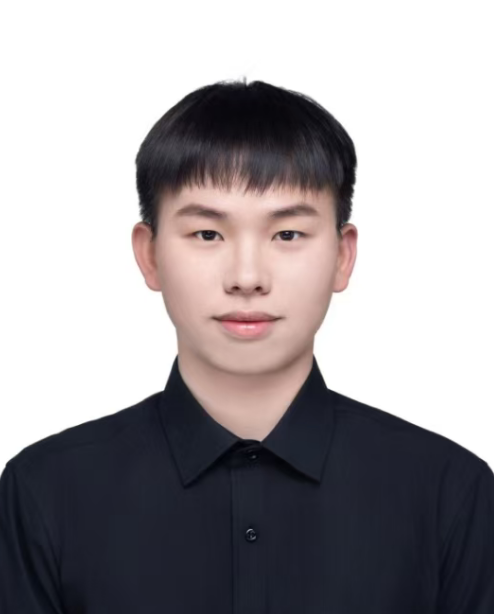}}]
{Pinhan Fu} is currently a Ph.D. student at the School of Computer Science, Wuhan University, Wuhan, China. He received a master's degree from Shanxi University, China, in 2025. His research interests include embodied intelligence, vision-language-action models, multimodal learning, and computer vision.
\end{IEEEbiography}

\vspace{-40pt}
\begin{IEEEbiography}
[{\bioimg{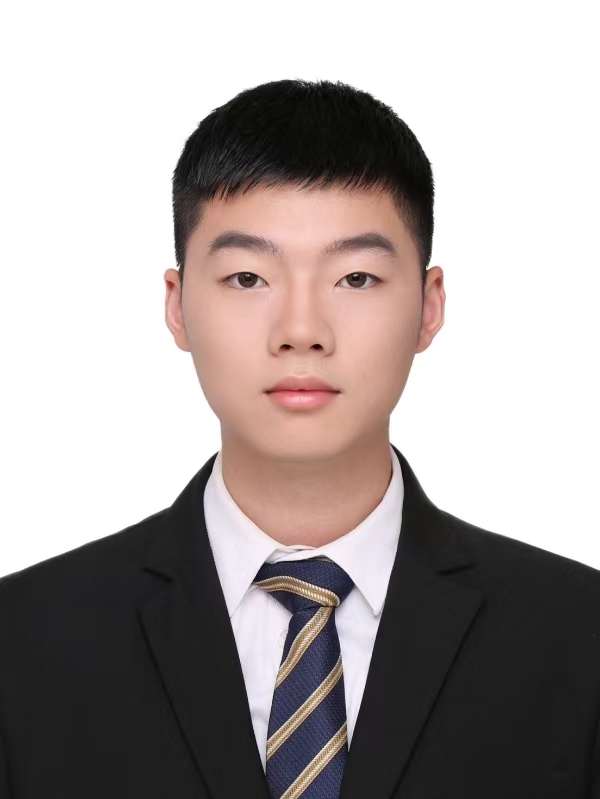}}]
{Ruilin Wang} received the B.Eng. degree in Automation from the Harbin Institute of Technology, Shenzhen, in 2025. He is currently pursuing a master's degree at the Institute of Automation, Chinese Academy of Sciences. His research interests lie in computer vision and robotics, including stereo vision, depth estimation, visual perception, and robot control.
\end{IEEEbiography}

\vspace{-40pt}
\begin{IEEEbiography}
[{\bioimg{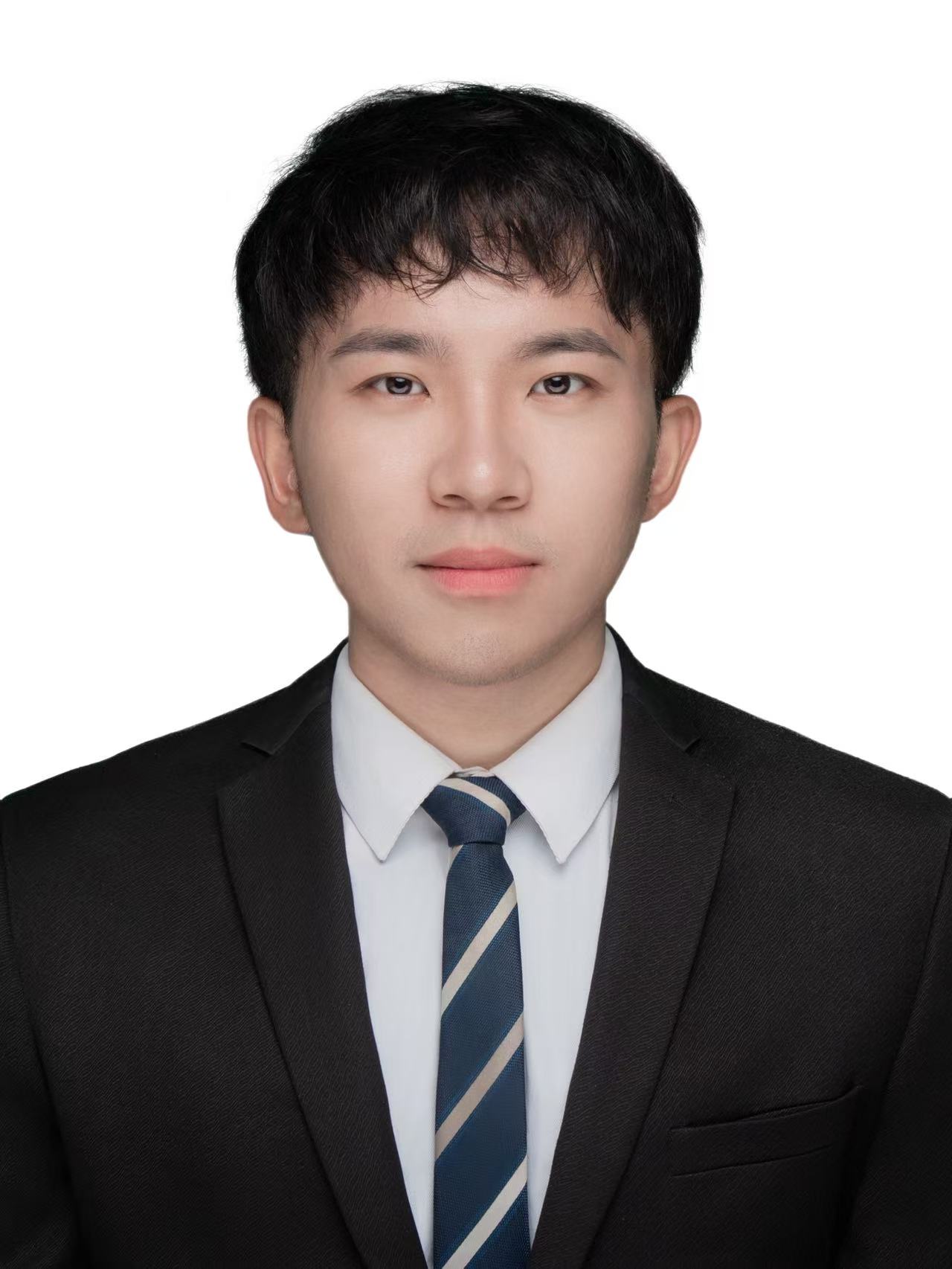}}]
{Wenke Huang} received the B.E. degree from the School of Computer Science, Wuhan University, Wuhan, China, in 2021, and the Ph.D. degree from Wuhan University. He is currently a Postdoctoral Research Fellow at Nanyang Technological University, Singapore. His research interests include federated learning and multimodal large language models.
\end{IEEEbiography}

\vspace{-30pt}
\begin{IEEEbiography}
[{\bioimg{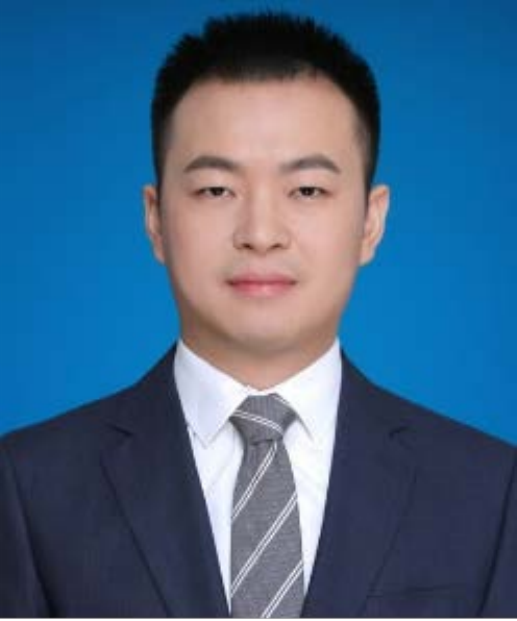}}]
{Mang Ye}(Senior Member, IEEE) received the Ph.D. degree in Computer Science from Hong Kong Baptist University in 2019. He was a Research Scientist at Inception Institute of Artificial Intelligence. He is currently a full professor with the School of Computer Science, Wuhan University, Wuhan, China. He has published more than 100 articles in top-tier venues. He serves as Area Chair of CVPR 2024, ECCV 2024, and NeurIPS 2024. He serves as Associate Editor for TIFS and TIP. His research interests focus on computer vision and federated learning.
\end{IEEEbiography}

\vspace{-30pt}
\begin{IEEEbiography}
[{\bioimg{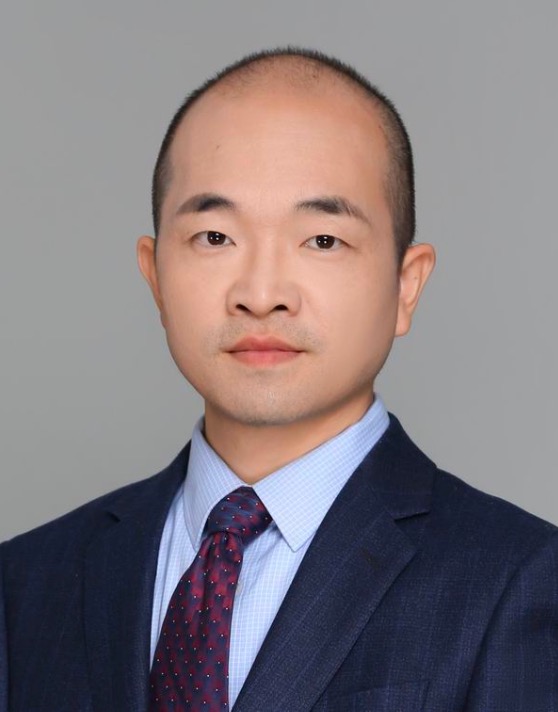}}]
{Qin Zou} (Senior Member, IEEE)  received his B.E. degree in Information Engineering and Ph.D. degree in Computer Vision from Wuhan University. From 2010 to 2011, he was a visiting Ph.D. student at the Computer Vision Lab, University of South Carolina. He is currently a professor with the School of Computer Science, Wuhan University. His research interests include machine vision, machine learning, and robotics.
\end{IEEEbiography}

\end{document}